\documentclass[10pt,twocolumn,letterpaper]{article}

\usepackage[datasets]{wacv}
\usepackage{multirow}
\usepackage{makecell}
\usepackage[absolute,overlay]{textpos}
\definecolor{wacvblue}{rgb}{0.21,0.49,0.74}
\usepackage[pagebackref,breaklinks,colorlinks,allcolors=wacvblue]{hyperref}
\def\wacvPaperID{1412} 
\def\confName{WACV}
\def\confYear{2027}

\title{
A Multi-Center Dataset and Benchmark for SYNTAX Segmentation of Coronary Angiograms
}

\author{%
Dominik Bernard Lau\textsuperscript{1,2}\thanks{Contact: dominik.lau@nask.pl}
\and
Jerzy Szyjut\textsuperscript{2,3}
\and
Hubert Malinowski\textsuperscript{2,4}
\and
Adam Brzeski\textsuperscript{2}
\and
Tomasz Dziubich\textsuperscript{2}
\and
Natalia Zieli\'nska\textsuperscript{5}
\and
Tomasz Figatowski\textsuperscript{5}
\and
Rados\l{}aw Targo\'nski\textsuperscript{5}
\\[4pt]
\textsuperscript{1}NASK National Research Institute, Poland \quad
\textsuperscript{2}Gda\'nsk University of Technology, Poland \quad \\
\textsuperscript{3}TU Wien, Austria \quad
\textsuperscript{4}Technical University of Munich, Germany \quad \\
\textsuperscript{5}Medical University of Gda\'nsk, Poland
}

\begin{document}
\maketitle
\begin{abstract}
Dense pixel-level classification of coronary angiography is essential for automated cardiovascular diagnostics, yet the computer vision community lacks gold-standard protocols and robust datasets to tackle this topologically complex, heavily long-tailed domain. To bridge this gap, we introduce CARDIAG: a rigorous, multi-center benchmark dataset designed to evaluate deep learning architectures on fine-grained SYNTAX segment classification. Unlike prior efforts, CARDIAG systematically prevents patient- and center-level data leakage while providing an unprecedented suite of annotations: highly imbalanced multi-class labels, explicit catheter and stenosis masks, expert uncertainty maps, temporal sequence frames, and spatial metadata. Using CARDIAG, we conduct the most extensive architectural sweep in this domain, benchmarking 24 models spanning classic CNNs, Vision Transformers (ViTs), and State-Space Models (SSMs). Moving beyond naive overlap metrics, we evaluate topological preservation (clDice), vessel diameter error, and uncertainty error-detection. Our analysis reveals an insight: modern large-kernel CNNs (particularly ConvNeXt V2 with DeepLabV3+) still outperform recent ViTs and SSMs under a shared training recipe, establishing a $\mathit{Dice}=0.456$. By ensembling complementary models, we push the baseline to $\mathit{Dice}=0.479$. Through leave-one-center-out cross-validation and data efficiency analyses, we explore the current limits of out-of-distribution generalization. Ultimately, CARDIAG establishes a foundation for advancing medical computer vision for cardiology across many downstream tasks, including lesion detection, temporal frame interpolation, and 3D reconstruction.
\end{abstract}

\begin{figure}[!t]
\centerline{\includegraphics[width=\columnwidth]{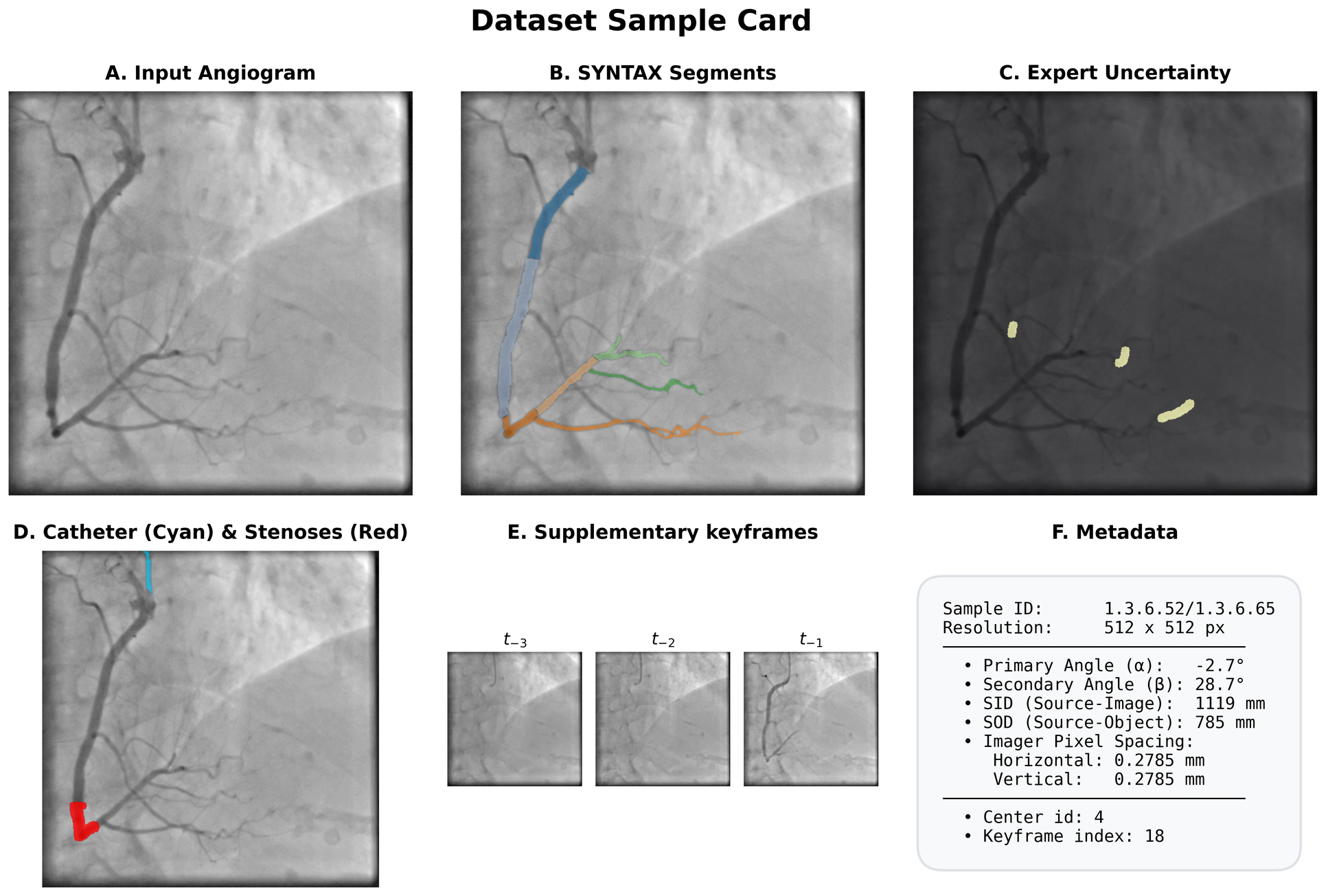}}
\caption{CARDIAG: A multi-label, multi-center dataset for the construction of coronary-angiography-applied models.}
\label{fig:dataset_card}
\end{figure}
\begin{textblock*}{540pt}(36pt,748pt)
{\scriptsize\raggedright
\copyright{} 2027 IEEE. Personal use of this material is permitted.  Permission from IEEE must be obtained for all other uses, in any current or future media, including reprinting/republishing this material for advertising or promotional purposes, creating new collective works, for resale or redistribution to servers or lists, or reuse of any copyrighted component of this work in other works.
\par}
\end{textblock*}

\section{Introduction}
\label{sec:introduction}

Segmenting coronary blood vessels in X-ray images to distinct classes related to their topological ancestry is a common first step towards automating the diagnosis of heart disease and measuring blockages \cite{Kaba2023TheAO}. 

We treat this segmentation as a highly imbalanced, dense pixel-level classification problem. Given an X-ray angiogram $X$, our goal is to train a model $f_\theta$ to predict the correct SYNTAX class \cite{SYNTAXtax} probability distribution for every single pixel $(i,j)$:
\begin{equation}
    \boldsymbol{\hat y(i,j)} = f_\theta(X, i, j) \in [0, 1]^{26}
    \label{eq:syntax_pd}
\end{equation}
including 25 SYNTAX classes and a catheter\footnote{Three rare SYNTAX segments (12b, 14a, 14b) have no annotated examples, so in the remainder of the paper metrics and figures are reported over the remaining 22 segments and the catheter, i.e., 23 classes in total.}.
This is a notoriously difficult task. Coronary arteries exhibit complex topologies, and the class distribution is heavily long tailed, driven by anatomical variation. Furthermore, the annotation process is bound by substantial aleatoric uncertainty from visual ambiguities and inter-observer disagreement. To make matters worse, the community lacks open, high-quality benchmarks to train and evaluate these models reliably.

While recent papers have made attempts at the task using Convolutional Neural Networks \cite{PPL, Jun2020}, Vision Transformers (ViTs) \cite{Xu2024}, and State-Space Models (SSMs) \cite{Zhang2026}, almost all of them rely on in-house datasets \cite{PPL, Park2023, Jun2020, Xian2020, Zhu2021, Chang2024, Zhang2026, Shi2020, Du2021, Zhai2019, AGMN, EAGMN, MGM, HAGMNUQ}, making their results difficult to reproduce. Furthermore, no prior work has conducted a unified, multi-family architectural sweep to establish definitive baselines for this specific task.

Even the current gold standard dataset, ARCADE \cite{ARCADE}, has potential flaws that make it risky for a reliable benchmark candidate; it lacks patient and examination IDs, meaning highly correlated frames from the same patient could accidentally leak across training and test splits. Because of this hypothetical data leakage, and because it provides no acquisition-site or device labels, we cannot reliably use it to test how well models generalize to new, out-of-distribution data.

We introduce \textbf{CARDIAG}, a multi-center dataset and an evaluation benchmark, specifically addressing the above issues. We use data from five medical centers and carefully structure it to guarantee no data leakage between patients or hospitals.  We also include explicit catheter masks, background frames, stenosis masks, temporal supplementary frames, DICOM acquisition metadata, and expert uncertainty maps. With these  annotations, the community can build systems that go beyond basic segmentation to tackle uncertainty quantification, frame interpolation, and 3D reconstruction. Our main contributions are summarized as follows:
\begin{itemize}
    \item \textbf{A Multi-Label, Multi-Center Dataset}: We release an exhaustively annotated benchmark that allows active prevention of data leakage, thus enabling more reliable studies of generalization capabilities of constructed models. By extending the standard multi-class labels with uncertainty masks, stenoses, supplementary temporal frames and metadata (such as acquisition configuration parameters), we permit a potential extension to a broad range of tasks aimed at the automation of common cardiological diagnostic procedures. 
    \item \textbf{A Wide Architectural Benchmark}: We rigorously evaluate 24 state-of-the-art vision models spanning legacy CNNs, large-kernel CNNs, ViTs, and SSMs to establish strong baselines for vascular segmentation across standard overlap metrics, topology-aware scores and vascular diameter truthfulness.

    \item \textbf{Generalization, Scaling, and Clinical Analysis:} We investigate real-world deployment limits through Leave-One-Center-Out Cross-Validation and data scaling curves to investigate hospital domain shift and data bottlenecks. In supplementary material, we provide an extensive meta-analysis correlating performance with patient demographics and C-arm projection geometry.
\end{itemize}

We release the benchmark source code at \url{https://github.com/cvlab-ai/cardiag-benchmark} and the dataset at \url{https://zenodo.org/records/19958730}.

\section{Related Works}
\label{sec:related_works}

\paragraph{Vascular Segmentation Architectures}
Most prior efforts in coronary angiography segmentation focus on coarse objectives, such as binary vessel extraction or identifying only the three major branches (them being LAD, RCA, and LCX). For these tasks, standard Convolutional Neural Networks (CNNs) already perform exceptionally well. For instance, Zhang \etal \cite{PPL} use a Progressive Perception Learning framework to utilize multi-scale features, while others have successfully applied ensembling \cite{Park2023} and U-Net variations like T-Net \cite{Jun2020}. For basic main-branch segmentation, even a simple U-Net can exceed a Dice score of 0.80 \cite{Xian2020}. 

For \textbf{binary segmentation}, researchers have deployed architectures like PSPNet \cite{Zhu2021}, Vision Transformers (ViTs) \cite{Xu2024}, and recently, State-Space Models (SSMs) such as HR-UMamba++ \cite{Zhang2026}. However, despite the known capability of SSMs to handle long-range dependencies, they have not yet been rigorously benchmarked against CNNs on the much more difficult SYNTAX segmentation task. 

Moving to \textbf{dense multi-class SYNTAX segmentation}, the literature remains highly fragmented with very few approaches advancing the domain in recent years. Techniques developed for the ARCADE challenge utilize ensembling \cite{ARCADE_Bilal}, multi-step pipelines split by vessel side \cite{ARCADE_Ku}, or vessel-map guided YOLOv8 \cite{ARCADE_Liu}. Other researchers have explored generative models, such as cGANs \cite{Yang2026} and UENet \cite{Shi2020}, though these require pre-computed binary masks, limiting their end-to-end utility. 

A separate subset of methods treats the arteries as graphs \cite{AGMN, EAGMN, MGM}. While models like HAGMN-UQ \cite{HAGMNUQ} provide greater inductive biases (and even attempt uncertainty quantification), their X-ray-to-graph extraction pipelines are susceptible to downstream error propagation and are ultimately non-robust to heavy artery entanglement. Crucially, almost all of these studies propose a single architectural approach (usually a U-Net or YOLO variant) rather than conducting a comprehensive, multi-family architectural sweep to establish true reproducible domain baselines.

\paragraph{Public Datasets of Coronary Angiography}
The most significant bottleneck in SYNTAX segmentation is the scarcity of public data. Many of the most notable high-performing approaches \cite{Du2021, Zhai2019} were trained on massive in-house datasets (ranging from 7,000 to 12,000 samples), putting their reproduction entirely out of reach for typical academic budgets (as the collection of labels of similar magnitude would require tremendous time investments from experts). 

Currently, the already-mentioned ARCADE dataset \cite{ARCADE} serves as the public gold standard for SYNTAX segmentation, containing 1500 labeled images collected at a single center in Almaty, Kazakhstan, using two different angiography systems. However, as noted by Seo \etal \cite{SeoSum_DiffusionBased_MICCAI2025}, there are concerns about the label quality. Furthermore, the authors attach neither the metadata nor patient or examination IDs, and as they suggest, there might be from 1 to 12 frames of the same patient present. No actions aimed at accounting for this potential leakage have been disclosed, which is a major flaw that makes the results of benchmarks carried out on this particular data not entirely reliable. 

Other public datasets exist such as XCAD \cite{Ma2021} and DCA1 \cite{CervantesSanchez2019} for binary segmentation, or CoronaryDominance \cite{Kruzhilov2025} and CardioSyntax \cite{CardioSyntax} for clinical scoring, but none provide labels for SYNTAX segmentation per se. The proposed CARDIAG benchmark directly addresses these issues by providing many axes of labels alongside comprehensive acquisition metadata including anonymized patient and center identifiers.

\section{CARDIAG Dataset}
\label{sec:dataset}

\paragraph{Annotation Procedure}
The labeling process was conducted between 2019 and 2021 by three specialists in interventional cardiology, each with at least three years of clinical experience, using a custom in-house annotation tool. Following SYNTAX score definitions \cite{SYNTAXtax}, annotations were performed on single angiographic frames with optimal contrast opacification. Angiographic images were obtained from anonymized examinations of 106 patients (median age 65 years, IQR 61--73): 60.4\% male (aged 36--87), 38.7\% female (aged 49--91), and one patient whose source DICOM records a placeholder birth date and unspecified sex. Acquisitions were carried out using an Axiom Artis system (Siemens Healthineers, Erlangen, Germany) across multiple European cardiology centers. Annotations were initially generated semi-automatically and subsequently refined manually. Prior to labeling, all experts underwent software training and consensus meetings to harmonize boundary definitions and annotation criteria.

\begin{figure}[!t]
\centerline{\includegraphics[width=\columnwidth]{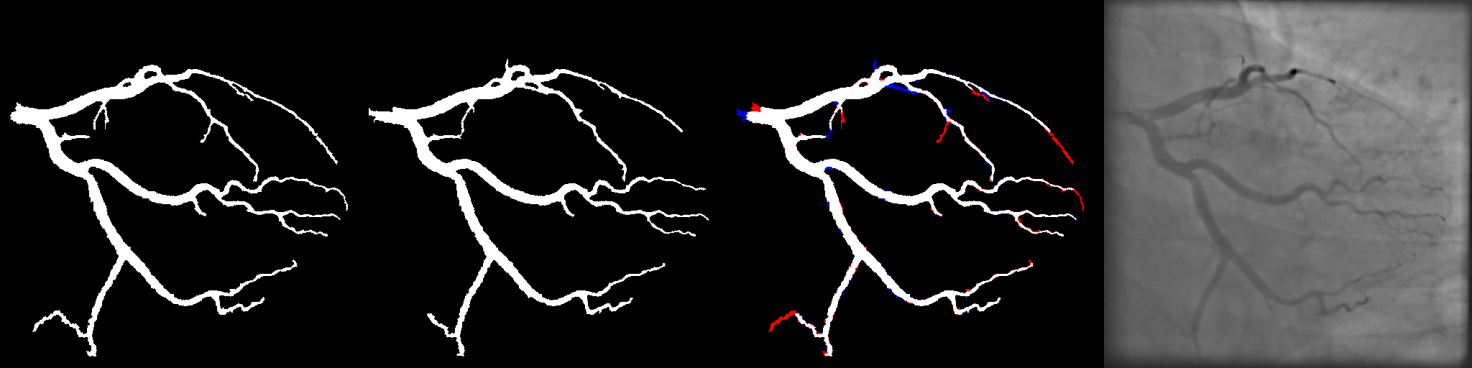}}
\centerline{\includegraphics[width=\columnwidth]{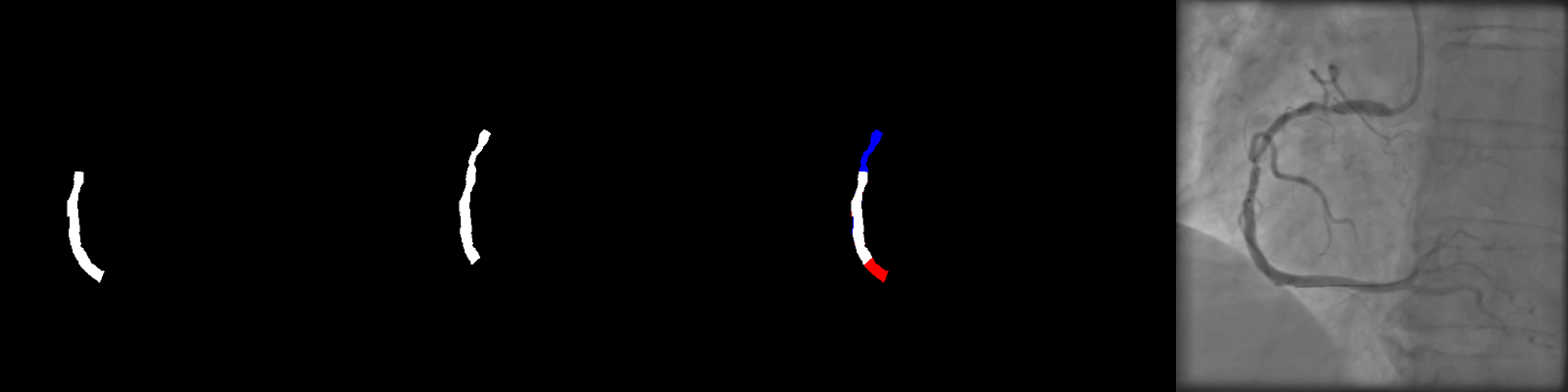}}
\caption{Example inter-observer comparison: (top row) full vascular tree segmentation and (bottom row) segment-specific annotation (RCA mid). Panels show annotations from Expert~1, Expert~2, their discrepancy map (white: overlap, red/blue: rater-specific markings), and the raw input frame.}
\label{fig:annotations}
\end{figure}

To evaluate \textbf{annotation quality and consistency}, validation was conducted on a randomly selected subset of 10 cases (\cref{fig:annotations}). For full vascular tree segmentation, inter-observer agreement was assessed via $Dice$ against a reference standard established by an expert with over 15 years of experience. The mean baseline $Dice$ was 0.87, reflecting substantial agreement. The most frequent discrepancies were observed at the distal ends of vessels, where one expert tended to exclude terminal branches considered clinically non-significant, while the other provided more detailed delineation of these structures.

To address this variance, we introduced an uncertainty-aware protocol: experts used \textbf{uncertainty masks} to coarsely mark ambiguous or clinically irrelevant vessel fragments. Excluding these regions from evaluation raised the mean $Dice$ to 0.9072. For stenosis labeling consistency, Fleiss' kappa reached $\kappa = 0.7611$, confirming substantial inter-rater reliability.

\begin{figure}[!t]
\centerline{\includegraphics[width=\columnwidth]{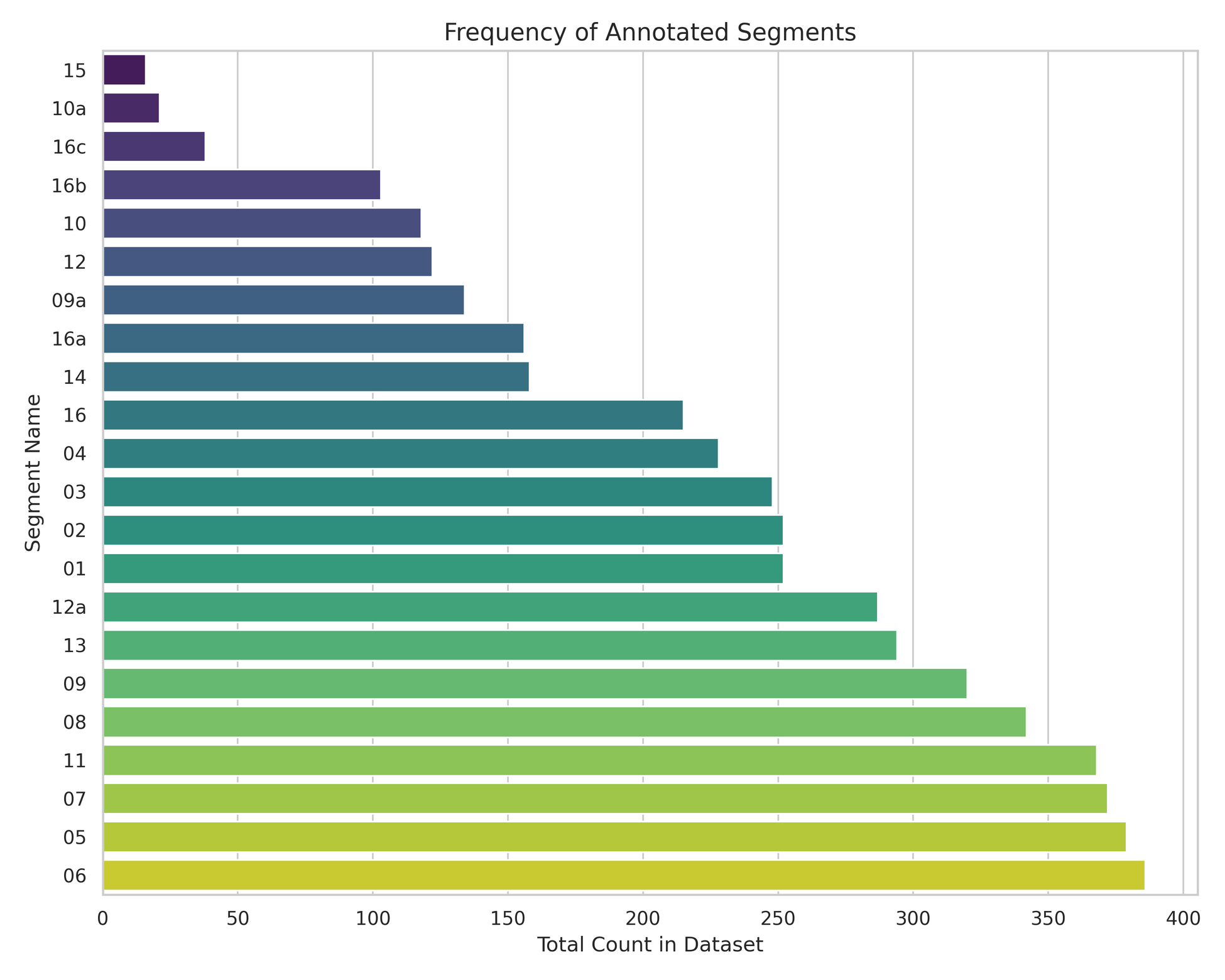}}
\caption{Frequency distribution of SYNTAX segments across the benchmark, highlighting long-tailed class imbalance.}
\label{fig:segment_counts}
\end{figure}

\paragraph{Label Composition}
Each sample in the CARDIAG benchmark consists of a primary X-ray frame accompanied by:
    (1) \textbf{Acquisition Metadata:} Standard DICOM parameters \cite{dicom}: Source-Image Distance (SID), Source-Object Distance (SOD), pixel spacing (mm), and primary/secondary C-arm angles ($\alpha, \beta$) enabling the reconstruction of detector plane position. Anonymized examination and sample IDs (\texttt{<exam\_id>/<proj\_id>}) are included to enable leakage-free stratified splits.
    (2) \textbf{Binary Vessel Masks:} Pixel-level foreground masks delineating the complete visible coronary tree.
    (3) \textbf{Multi-Class SYNTAX Masks:} Fine-grained binary masks per individual SYNTAX anatomical segment (\eg, \texttt{04.png}).
    (4) \textbf{Catheter Masks:} Dedicated binary masks isolating the contrast-injection catheter to prevent false-positive vessel detections.
    (5) \textbf{Stenosis Masks}: Binary masks marking narrowed vessels (\texttt{narrowings.png}) for potential stenosis detection and grading tasks.
    (6) \textbf{Temporal Sequences:} Static background acquisitions (\texttt{background.png}), along with up to 52 preceding and 80 subsequent intermediate frames per projection sequence.
    (7) \textbf{Uncertainty Maps:} Masks marking ambiguous distal vascular territory for possible uncertainty quantification.

As shown in \cref{fig:segment_counts}, the segment frequencies exhibit a severe long-tailed distribution, with distal vessels appearing substantially less frequently, a critical challenge for model convergence. Further demographic and anatomical distributions are detailed in the supplementary material.

The final benchmark comprises 645 fully annotated primary frames of $512\times512$ pixels from 106 patients and five centers, accompanied by 29{,}056 temporal frames from the same acquisition sequences. Beyond dense SYNTAX and catheter segmentation, CARDIAG could support further downstream vision tasks, which we do not benchmark here, including: (1) \textbf{stenosis detection}, (2) \textbf{temporal frame interpolation} (via multi-frame context), and (3) \textbf{3D vascular reconstruction} leveraging calibrated C-arm geometry ($\alpha, \beta, \text{SID}, \text{SOD}$). All distributed data are stripped of demographic identifiers to prevent patient re-identification.

\section{Methodology \& Evaluation}

\subsection{Models}

To rigorously evaluate the CARDIAG benchmark, we select 24 architectures spanning five identified groups, each representing different architectural design choices for semantic segmentation:

\begin{itemize}[leftmargin=*, topsep=0pt]
    \item \textbf{Standard CNNs:} Including U-Net \cite{unet}, Attention U-Net \cite{oktay2018attention}, PSPNet \cite{PSPNet}, FPN \cite{FPN}, UPerNet \cite{UPERNET}, and DeepLabV3/V3+ \cite{DEEPLABV3, DEEPLABV3PLUS}. These models rely on the strong inductive bias of local pixel correlation and satisfy translation equivariance \cite{geodl}, making them well-established baselines for dense medical segmentation.
    \item \textbf{Vision Transformers (ViTs):} Including SegFormer \cite{segformer} and SwinUNETR \cite{swinunetr}. These architectures capture global dependencies via self-attention. To mitigate the lack of inherent spatial priors found in CNNs, they employ structural adaptations such as shifted windows or hierarchical patch merging coupled with convolutional feature extraction.
    \item \textbf{Modern \& Large-Kernel CNNs:} Including ConvNeXt V2 \cite{convnext2}, MedNeXt \cite{mednext}, SegNeXt \cite{segnext}, and RepLKNet \cite{replknet}, utilizing DeepLab or Hamburger decoding heads. These architectures seek to bridge the gap between paradigms, combining the local priors of standard CNNs with the massive receptive fields of transformers.
    \item \textbf{State-Space Models (SSMs):} Including VM-UNet \cite{VMUnet}, Swin-UMamba \cite{swinumamba} and Mamba-UNet \cite{mambaunet}. These models address the quadratic computational bottleneck of ViT self-attention by leveraging linear-complexity sequence modeling. They are typically hybridized with CNN components to retain local spatial priors while efficiently processing global context.
    \item \textbf{Domain-Pretrained Baselines:} Including XRayVision \cite{xrayvision} (with a U-Net decoder) and RadImageNet \cite{radimagenet} (with DeepLab). These test the hypothesis that initializations learned from massive, domain-specific medical datasets (\eg, chest X-rays and multi-modal radiology scans) provide more effective representations for angiographic textures than standard ImageNet pretraining.
\end{itemize}

To obtain a stronger baseline, we construct a heterogeneous ensemble combining three architecturally diverse models: ConvNeXt V2, FPN, and Mamba-UNet. We aggregate the network outputs using three distinct pixel-level voting strategies: 
(1) \textbf{Hard voting}, which applies a simple majority rule across discrete class predictions; 
(2) \textbf{Soft voting}, which computes the unweighted mean of the predicted class probabilities across all models; and 
(3) \textbf{Entropy-weighted voting}, which leverages the uncertainty of the individual architectures. In this strategy, the final prediction is a weighted average where each model's contribution is inversely proportional to the entropy of its output distribution. This mechanism dynamically assigns greater influence to the network exhibiting the highest confidence for any given pixel.


\textbf{Training:}
To ensure a fair comparison, all architectures are trained under a unified optimization protocol. The overall objective function is a combination of multi-class Cross-Entropy and Dice loss:
\begin{equation}
    \mathcal{L} = \mathcal{L}_{CE} + \mathcal{L}_{Dice}
    \label{eq:loss_function}
\end{equation}
We optimize the networks using AdamW with an initial learning rate $\eta=10^{-4}$, weight decay $\lambda=10^{-2}$, and momentum parameters $\beta_1 = 0.5, \beta_2 = 0.999$. The learning rate is modulated by a cosine annealing scheduler with warm restarts ($T_0=20$, $T_{mult}=1$, $\eta_{min}=10^{-6}$). These hyperparameters were established via pilot experiments to ensure stable convergence across all model families and are kept constant across models. This common protocol provides a ranking under one recipe; it does not guarantee that the recipe is equally well suited to every architecture, therefore the differences may partly reflect training-recipe fit rather than architecture alone.

To further stabilize the optimization landscape we employ an Exponential Moving Average (EMA) of the model weights \cite{Izmailov2018AveragingWL}. At each training step $t$, while the active parameters $\theta_t$ are updated via backpropagation, the EMA weights $\theta_t^{EMA}$ are updated according to:
\begin{equation}
    \theta_t^{EMA} = \alpha_t \theta_{t-1}^{EMA} + (1-\alpha_t) \theta_t
    \label{eq:ema}
 \end{equation}
We apply a step-dependent decay rate $\alpha_t = \min(\alpha_{max}, 1-1/t)$ with $\alpha_{max} = 0.999$. In this context, EMA acts as a temporal ensemble, smoothing optimization trajectories and guiding models toward broader, more generalizable minima without introducing additional inference overhead. All models are trained for a maximum of 700 epochs, utilizing an early stopping mechanism with a patience of 100 epochs monitored on the validation set. Details on the model input, augmentation and output channels are given in supplementary material.

\subsection{Evaluation Metrics}
To comprehensively evaluate model performance across both pixel-level accuracy and clinical utility, we define a suite of metrics.

\textbf{Overlap and Clinical Importance} 
As a baseline, we compute the macro-averaged Dice score across all valid vessel classes, denoted $\mathit{Dice}^{(m)}$ (computed per frame over the foreground classes and then averaged over frames; a class that is absent from both the ground truth and the prediction is undefined and excluded from the average, whereas a class that is present in only one of them scores 0). However, standard overlap metrics treat all segments equally, which does not reflect clinical reality. Therefore, we introduce importance-weighted Dice subsets based on the SYNTAX nomenclature \cite{SYNTAXtax}:
\begin{itemize}[leftmargin=*, topsep=0pt]
    \item Critical ($\mathit{Dice}^{(ci)}$): Segments directly impacting patient survival (1 [RCA proximal], 5 [Left main], 6 [LAD proximal], 11 [Proximal circumflex]).
    \item High ($\mathit{Dice}^{(hi)}$): Major branches (7 [LAD mid], 2 [RCA mid], 9 [First diagonal], 12 [LCX intermediate/anterolateral], 13 [Distal circumflex], 14 [Left posterolateral], 3 [RCA distal]).
    \item Low ($\mathit{Dice}^{(li)}$): All remaining distal branches.
\end{itemize}

\textbf{Spatial and Topological Accuracy} 
To penalize boundary inaccuracies, we calculate the 95th percentile Hausdorff Distance ($HD_{95}$):
\begin{equation}
    HD_{95}(P, G) = \max\{h_{95}(P, G),\, h_{95}(G, P)\}
    \label{eq:hd_95}
\end{equation}
where $h_{95}(A, B)$ is the 95th percentile of the Euclidean distances (in pixels) from the boundary points of mask $A$ to the nearest boundary point of mask $B$, and $P$, $G$ are the predicted and ground-truth masks. It is computed per class and averaged over the classes for which it is defined. Because overlap and distance metrics fail to capture the structural connectivity of vascular trees, we also report $\mathit{clDice}$ \cite{Shit2021}, which explicitly measures topological preservation:
\begin{equation}
    \mathit{clDice} = \frac{2 \cdot T_{\text{prec}}(S_P, V_G) \cdot T_{\text{sens}}(S_G, V_P)}{T_{\text{prec}}(S_P, V_G) + T_{\text{sens}}(S_G, V_P)}
    \label{eq:cldice}
\end{equation}
where $V_P, V_G$ denote the predicted and ground truth volumes, and $S_P, S_G$ represent their respective topological skeletons. Topological precision and sensitivity are defined as $T_{\text{prec}} = \frac{|S_P \cap V_G|}{|S_P|}$ and $T_{\text{sens}} = \frac{|S_G \cap V_P|}{|S_G|}$.

\textbf{Diameter Error and Error-Detection}
To determine if a model accurately reflects vessel width (which is a prerequisite for stenosis grading), we propose the \textbf{Diameter Error} ($\boldsymbol{DE}$):
\begin{equation}
    DE = \frac{1}{|S_G|} \sum_{x \in S_G} |D_P(x) - D_G(x)|
    \label{eq:de}
\end{equation}
where $D_P(x)$ and $D_G(x)$ represent the vessel diameter (in pixels) at the ground-truth skeleton pixel $x$, derived via a distance transform.

Finally, we assess how well the model's uncertainty $U$ detects its actual mistakes $E$. We estimate pixel-wise uncertainty $U$ as the entropy of the class distribution averaged over 5 stochastic forward passes with Monte Carlo (MC) dropout (if available, otherwise we use the entropy of softmax output or, for ensembles, the entropy of the member probabilities). By treating uncertainty $U$ as the predicted probability of a classification error ($P(\hat{y} \neq y) = U$), we define the Error-Detection AUROC ($\mathcal{C}$) as the Area Under the Receiver Operating Characteristic curve:
\begin{equation}
    \mathcal{C} = \text{AUC}(U, E)
    \label{eq:ce}
\end{equation}
Model computational complexity is reported in GFLOPs.

\subsection{Experimental Setup}

\begin{table*}[!th]
\scriptsize
\caption{Quantitative benchmark results across all 24 architectures. Rows are grouped by family (standard CNNs, modern CNNs, ViTs, SSMs, domain-pretrained) and sorted by $\mathit{Dice}^{(m)}$ within each family. The best result in each column is bolded and the second best is underlined. Values are reported as mean $\pm$ the 95\% bootstrap confidence interval.}
\label{tab:results_table}
\setlength{\tabcolsep}{3pt}
\begin{tabular*}{\textwidth}{@{\extracolsep{\fill}} l r r r r r r r r @{}}
\hline
Model & $\mathit{Dice}^{(m)}$ & $\mathit{Dice}^{(ci)}$ & $\mathit{Dice}^{(hi)}$ & $\mathit{Dice}^{(li)}$ & $DE$ & $clDice$ & $\mathcal{C}$ & $HD_{95}$ \\
\hline
FPN & $0.437_{{\pm 0.025}}$ & $0.668_{{\pm 0.039}}$ & $0.428_{{\pm 0.042}}$ & $0.266_{{\pm 0.030}}$ & $2.463_{{\pm 0.125}}$ & $0.800_{{\pm 0.017}}$ & $0.973_{{\pm 0.005}}$ & $56.904_{{\pm 4.852}}$ \\
DeepLab V3+ & $0.432_{{\pm 0.024}}$ & $0.680_{{\pm 0.034}}$ & $0.429_{{\pm 0.043}}$ & $0.242_{{\pm 0.028}}$ & $2.537_{{\pm 0.114}}$ & $0.789_{{\pm 0.016}}$ & $0.963_{{\pm 0.005}}$ & $57.686_{{\pm 4.898}}$ \\
DeepLab & $0.417_{{\pm 0.025}}$ & $0.671_{{\pm 0.037}}$ & $0.427_{{\pm 0.043}}$ & $0.236_{{\pm 0.028}}$ & $2.551_{{\pm 0.109}}$ & $0.799_{{\pm 0.016}}$ & $0.937_{{\pm 0.007}}$ & $59.754_{{\pm 5.203}}$ \\
Attention UNet & $0.396_{{\pm 0.025}}$ & $0.650_{{\pm 0.036}}$ & $0.422_{{\pm 0.045}}$ & $0.204_{{\pm 0.027}}$ & $2.442_{{\pm 0.126}}$ & $0.778_{{\pm 0.017}}$ & $0.956_{{\pm 0.006}}$ & $89.172_{{\pm 7.408}}$ \\
PSPNet & $0.362_{{\pm 0.024}}$ & $0.591_{{\pm 0.038}}$ & $0.361_{{\pm 0.043}}$ & $0.194_{{\pm 0.024}}$ & $3.271_{{\pm 0.115}}$ & $0.724_{{\pm 0.020}}$ & $0.977_{{\pm 0.003}}$ & $62.956_{{\pm 5.024}}$ \\
UPerNet & $0.356_{{\pm 0.023}}$ & $0.609_{{\pm 0.041}}$ & $0.361_{{\pm 0.044}}$ & $0.185_{{\pm 0.023}}$ & $2.680_{{\pm 0.140}}$ & $0.759_{{\pm 0.018}}$ & $0.973_{{\pm 0.005}}$ & $68.974_{{\pm 5.223}}$ \\
UNet & $0.350_{{\pm 0.022}}$ & $0.599_{{\pm 0.037}}$ & $0.342_{{\pm 0.038}}$ & $0.163_{{\pm 0.020}}$ & $2.784_{{\pm 0.109}}$ & $0.751_{{\pm 0.014}}$ & $0.942_{{\pm 0.007}}$ & $94.775_{{\pm 6.612}}$ \\
\hline
ConvNeXt V2 DeepLab & $\underline{0.456}_{{\pm 0.027}}$ & $\underline{0.694}_{{\pm 0.038}}$ & $\underline{0.464}_{{\pm 0.045}}$ & $\underline{0.273}_{{\pm 0.031}}$ & $2.332_{{\pm 0.115}}$ & $\mathbf{0.810}_{{\pm 0.016}}$ & $0.959_{{\pm 0.006}}$ & $57.662_{{\pm 5.659}}$ \\
RepLKNet Hamburger & $0.425_{{\pm 0.024}}$ & $0.660_{{\pm 0.037}}$ & $0.430_{{\pm 0.044}}$ & $0.238_{{\pm 0.028}}$ & $2.930_{{\pm 0.129}}$ & $0.750_{{\pm 0.019}}$ & $0.958_{{\pm 0.005}}$ & $54.428_{{\pm 4.756}}$ \\
ConvNeXt FPN & $0.424_{{\pm 0.023}}$ & $0.651_{{\pm 0.034}}$ & $0.426_{{\pm 0.042}}$ & $0.237_{{\pm 0.027}}$ & $2.975_{{\pm 0.096}}$ & $0.769_{{\pm 0.017}}$ & $0.975_{{\pm 0.003}}$ & $\underline{52.108}_{{\pm 4.749}}$ \\
ConvNeXt DeepLab & $0.423_{{\pm 0.026}}$ & $0.681_{{\pm 0.035}}$ & $0.424_{{\pm 0.044}}$ & $0.240_{{\pm 0.027}}$ & $2.334_{{\pm 0.112}}$ & $\mathbf{0.810}_{{\pm 0.017}}$ & $0.941_{{\pm 0.008}}$ & $61.436_{{\pm 5.723}}$ \\
MedNeXt & $0.404_{{\pm 0.024}}$ & $0.669_{{\pm 0.036}}$ & $0.430_{{\pm 0.045}}$ & $0.213_{{\pm 0.025}}$ & $2.384_{{\pm 0.119}}$ & $0.791_{{\pm 0.017}}$ & $0.957_{{\pm 0.006}}$ & $68.186_{{\pm 6.060}}$ \\
ConvNeXt V2 Hamburger & $0.397_{{\pm 0.023}}$ & $0.642_{{\pm 0.036}}$ & $0.406_{{\pm 0.043}}$ & $0.217_{{\pm 0.026}}$ & $3.061_{{\pm 0.123}}$ & $0.742_{{\pm 0.018}}$ & $0.972_{{\pm 0.003}}$ & $59.630_{{\pm 5.015}}$ \\
SegNeXt & $0.386_{{\pm 0.023}}$ & $0.644_{{\pm 0.037}}$ & $0.407_{{\pm 0.043}}$ & $0.208_{{\pm 0.024}}$ & $2.960_{{\pm 0.116}}$ & $0.760_{{\pm 0.017}}$ & $0.970_{{\pm 0.005}}$ & $61.459_{{\pm 4.929}}$ \\
ConvNeXt UPerNet & $0.315_{{\pm 0.022}}$ & $0.562_{{\pm 0.041}}$ & $0.331_{{\pm 0.040}}$ & $0.153_{{\pm 0.019}}$ & $3.031_{{\pm 0.151}}$ & $0.736_{{\pm 0.020}}$ & $0.963_{{\pm 0.009}}$ & $70.901_{{\pm 5.498}}$ \\
RepLKNet DeepLab & $0.280_{{\pm 0.017}}$ & $0.507_{{\pm 0.031}}$ & $0.298_{{\pm 0.035}}$ & $0.126_{{\pm 0.017}}$ & $5.742_{{\pm 0.153}}$ & $0.515_{{\pm 0.021}}$ & $0.948_{{\pm 0.004}}$ & $60.907_{{\pm 4.945}}$ \\
\hline
SegFormer (Big) & $0.428_{{\pm 0.025}}$ & $0.678_{{\pm 0.036}}$ & $0.442_{{\pm 0.045}}$ & $0.241_{{\pm 0.028}}$ & $2.681_{{\pm 0.130}}$ & $0.776_{{\pm 0.017}}$ & $0.971_{{\pm 0.005}}$ & $58.728_{{\pm 4.997}}$ \\
Segformer & $0.401_{{\pm 0.023}}$ & $0.638_{{\pm 0.035}}$ & $0.418_{{\pm 0.043}}$ & $0.212_{{\pm 0.024}}$ & $2.614_{{\pm 0.112}}$ & $0.783_{{\pm 0.016}}$ & $\underline{0.979}_{{\pm 0.003}}$ & $68.714_{{\pm 5.281}}$ \\
Swin-UNETR & $0.372_{{\pm 0.024}}$ & $0.603_{{\pm 0.040}}$ & $0.387_{{\pm 0.044}}$ & $0.188_{{\pm 0.024}}$ & $2.493_{{\pm 0.121}}$ & $0.763_{{\pm 0.017}}$ & $0.941_{{\pm 0.007}}$ & $77.336_{{\pm 6.416}}$ \\
\hline
VMUnet & $0.438_{{\pm 0.023}}$ & $0.674_{{\pm 0.034}}$ & $0.461_{{\pm 0.042}}$ & $0.255_{{\pm 0.028}}$ & $2.280_{{\pm 0.114}}$ & $\mathbf{0.810}_{{\pm 0.017}}$ & $0.978_{{\pm 0.003}}$ & $57.649_{{\pm 4.661}}$ \\
Swin-UMamba & $0.427_{{\pm 0.024}}$ & $0.678_{{\pm 0.035}}$ & $0.450_{{\pm 0.044}}$ & $0.237_{{\pm 0.027}}$ & $\underline{2.226}_{{\pm 0.117}}$ & $0.798_{{\pm 0.016}}$ & $0.960_{{\pm 0.005}}$ & $62.128_{{\pm 5.825}}$ \\
Mamba-UNet & $0.427_{{\pm 0.023}}$ & $0.683_{{\pm 0.034}}$ & $0.451_{{\pm 0.044}}$ & $0.241_{{\pm 0.025}}$ & $\mathbf{2.176}_{{\pm 0.111}}$ & $\underline{0.804}_{{\pm 0.016}}$ & $0.976_{{\pm 0.003}}$ & $59.702_{{\pm 5.307}}$ \\
\hline
Rad ImageNet DeepLab & $0.407_{{\pm 0.024}}$ & $0.652_{{\pm 0.037}}$ & $0.415_{{\pm 0.044}}$ & $0.228_{{\pm 0.028}}$ & $2.721_{{\pm 0.128}}$ & $0.765_{{\pm 0.018}}$ & $0.977_{{\pm 0.004}}$ & $61.415_{{\pm 4.929}}$ \\
XRay Vision UNet & $0.360_{{\pm 0.024}}$ & $0.602_{{\pm 0.041}}$ & $0.355_{{\pm 0.040}}$ & $0.199_{{\pm 0.024}}$ & $2.544_{{\pm 0.135}}$ & $0.784_{{\pm 0.017}}$ & $0.968_{{\pm 0.004}}$ & $71.179_{{\pm 5.629}}$ \\
\hline
Ensemble (Hard Voting) & $\mathbf{0.479}_{{\pm 0.026}}$ & $\mathbf{0.705}_{{\pm 0.036}}$ & $\mathbf{0.493}_{{\pm 0.045}}$ & $\mathbf{0.286}_{{\pm 0.031}}$ & $2.428_{{\pm 0.130}}$ & $0.801_{{\pm 0.018}}$ & $\mathbf{0.981}_{{\pm 0.003}}$ & $\mathbf{51.537}_{{\pm 5.022}}$ \\
\hline
\end{tabular*}

\end{table*}

\textbf{Splitting Strategy:}
The benchmark dataset is divided into a single training (70\%), validation (10\%), and test (20\%) fold. To prevent data leakage, we stratify these splits at both the institutional and patient levels. The data is first grouped by the originating medical center, and subsequently by individual patient IDs. Entire patients, rather than individual imaging frames, are randomly allocated into the respective splits. This dual-level stratification ensures that highly correlated frames from a single patient remain strictly isolated, while simultaneously preserving the proportional representation of different hospitals across the folds to mitigate center-specific bias. The per-center and per-split composition, including class coverage, is given in supplementary material and the list of samples constituting each split is provided alongside the dataset.

\textbf{Out-of-Distribution Generalization:}
To evaluate cross-center robustness, we perform a Leave-One-Center-Out Cross-Validation. For this experiment, we select the top-performing architecture from each model family. The models are trained on $N-1$ centers and evaluated on the held-out center. Training is conducted for 300 epochs, with all other hyperparameters matching the main benchmark protocol.

\textbf{Data Efficiency:}
To investigate whether the primary bottleneck in coronary segmentation is architectural capability or data scarcity, we conduct data efficiency scaling tests. We train the top-performing architectures on varying fractions of the training set: $tdf \in \{0.2, 0.4, 0.5, 0.8, 1.0\}$. To ensure a fair computational comparison regardless of dataset size, we standardize the training budget by fixing the maximum number of iterations at 8,000 (disabling early stopping and epochs) while maintaining identical optimization settings. All models are ultimately evaluated on the full, unmodified test set.

\begin{figure*}[!t]
\centering
\includegraphics[width=0.60\textwidth]{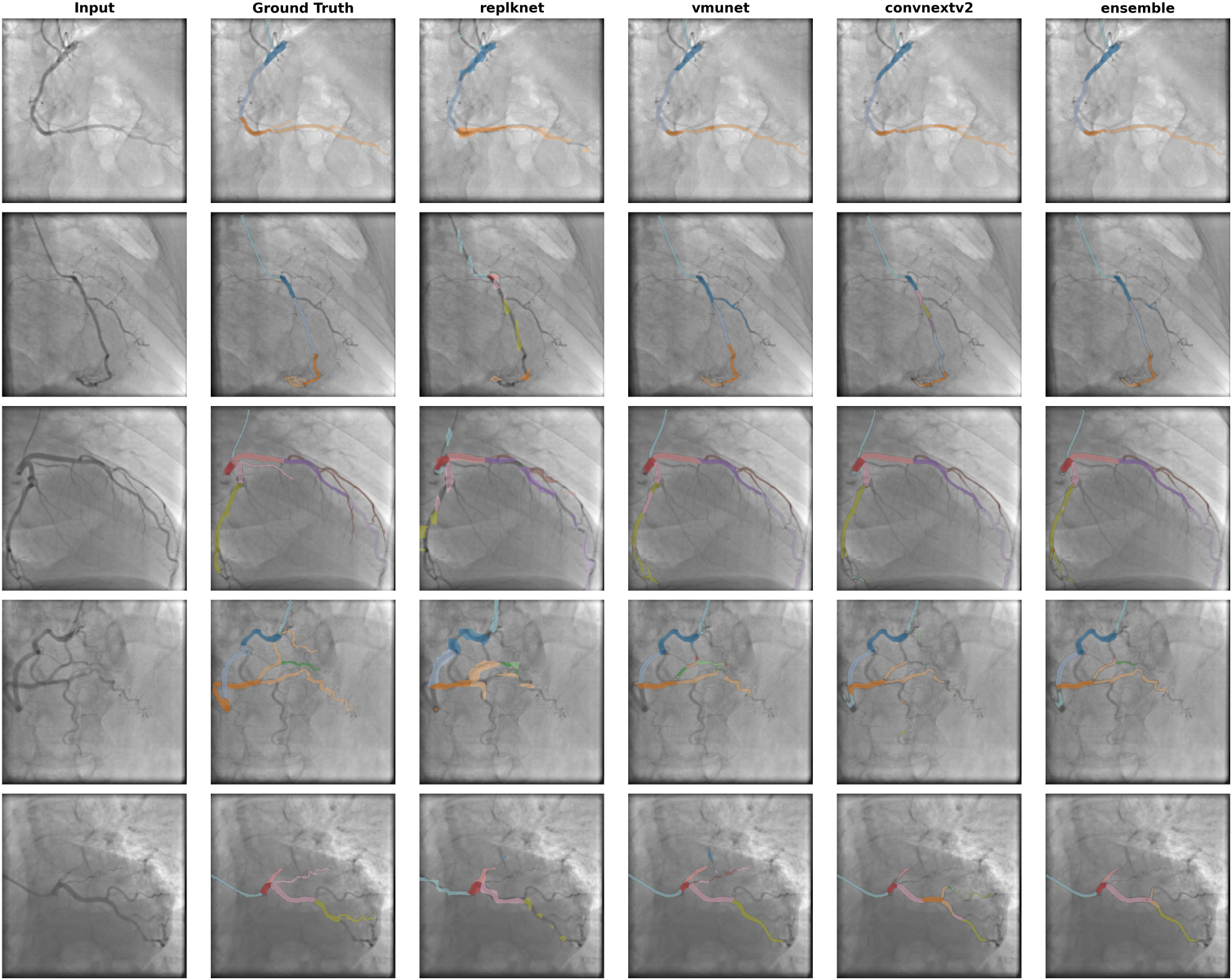}
\caption{Qualitative comparison of inference results across distinct architectural families. The visualizations highlight the failure of RepLKNet versus the fine-grained continuity preserved by ConvNeXt V2 and the Ensemble.}
\label{fig:qualitative_comparison}
\end{figure*}

\section{Benchmark Results \& Analysis}
\label{sec:results}

In this section, we present the findings of our multi-family model benchmark. We analyze absolute segmentation performance, investigate cross-center generalization and data efficiency. Additionally, in supplementary material we evaluate algorithmic bias with respect to demographic and acquisition parameters.

\subsection{Main Architectural Benchmark}

\textbf{CNNs vs. ViTs vs. SSMs:} 
As detailed in \cref{tab:results_table}, the combination of ConvNeXt V2 with a DeepLabV3+ decoder emerges as the strongest standalone architecture. It dominates the Dice categories ($\mathit{Dice}^{(m)}$, $\mathit{Dice}^{(ci)}$, $\mathit{Dice}^{(hi)}$) and ties for the best topological preservation ($\mathit{clDice}$). Under our shared training recipe, this model ranks above the transformer- and SSM-based models; translation equivariance and multi-scale aggregation (ASPP) is one of the possible contributing factors. We confirm this advantage is statistically significant via pairwise Wilcoxon signed-rank tests reported in the supplementary material.

While Vision Transformers (\eg, SegFormer) achieve competitive mean overlap scores, they generally fail to surpass the leading modern CNNs, which may relate to weaker local spatial priors. State-Space Models (VM-UNet, Swin-UMamba, Mamba-UNet) deliver highly respectable overlap results on par with CNN baselines like FPN. Crucially, while SSMs trail slightly in absolute overlap, \textbf{Mamba-UNet achieves the lowest Diameter Error ($\boldsymbol{DE} = 2.176$)}. This makes SSMs promising candidates for stenosis quantification (although stenosis assessment itself is not benchmarked here). In supplementary material we additionally show performance of each model across individual SYNTAX classes.

An ensemble of three diverse models (ConvNeXt V2, FPN, and Mamba-UNet) pushes $\mathit{Dice}^{(m)}$ to a peak of 0.479 and improves $HD_{95}$ (51.5 vs.\ 57.7), yet it does not improve topology or diameter over the best models - $\mathit{clDice}$ is 0.801 vs.\ 0.810 for ConvNeXt V2, and $DE$ worsens from 2.332 to 2.428 (Mamba-UNet remains best at 2.176).

\textbf{Architectural Incompatibilities \& Pre-training:}
Our sweep reveals that arbitrarily pairing modern encoders and decoders does not guarantee performance gains. The most extreme example is RepLKNet paired with DeepLab, which drastically underperforms even the baseline U-Net. A plausible explanation is an architectural clash: the exceptionally large kernels of RepLKNet conflict with the dilated convolutions of DeepLab, which would result in imprecise, low-resolution segmentation masks; swapping the decoder to Hamburger recovers performance, consistent with this explanation. 

Initializing networks with domain-specific medical weights (XRayVision or RadImageNet) yielded marginal or even negative improvements compared to standard ImageNet pre-training. This suggests that the visual textures of coronary angiography might not share substantial transferable traits with standard chest X-rays or multi-modal CT/MRI scans.

\begin{figure}[!t]
\centering
\includegraphics[width=0.9\columnwidth]{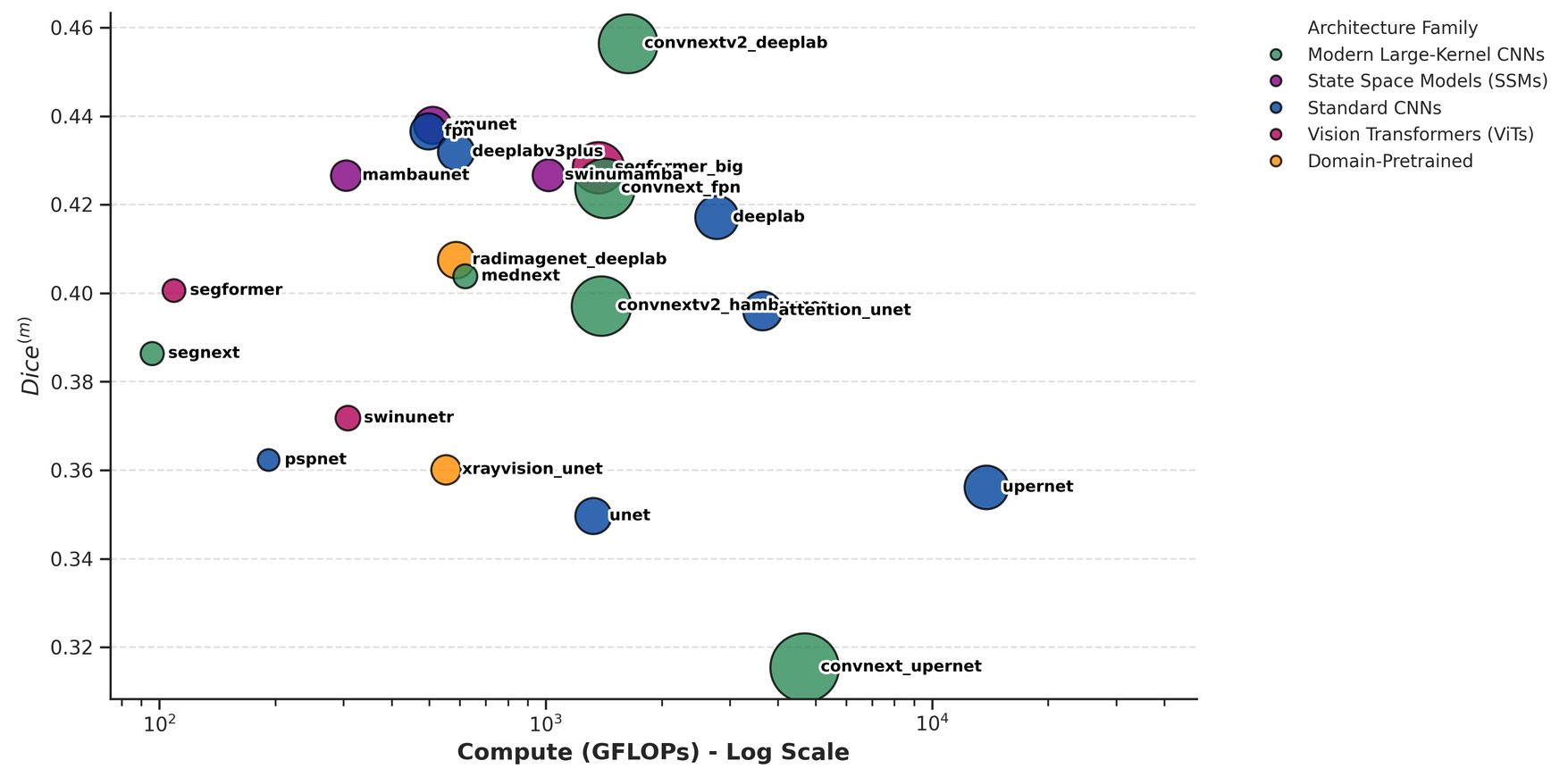}\\[-1pt]
\includegraphics[width=0.9\columnwidth]{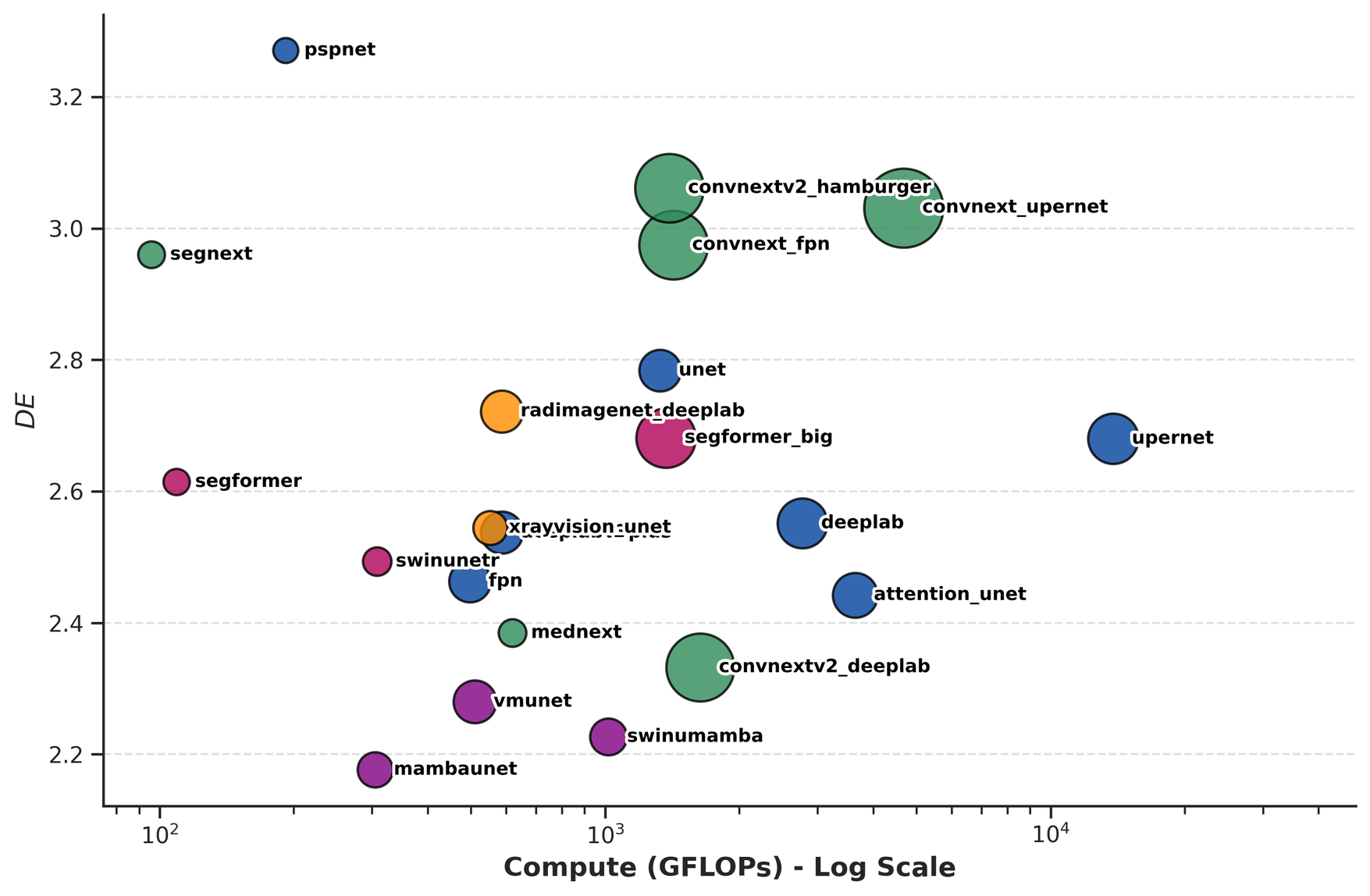}
\caption{Key performance metrics ($\mathit{Dice}^{(m)}$ and $DE$) relative to computational cost (GFLOPs). Note the logarithmic scale of the X-axis.}
\label{fig:flops_f1}
\end{figure}

\begin{figure}[!t]
\centering
\includegraphics[width=1.0\columnwidth]{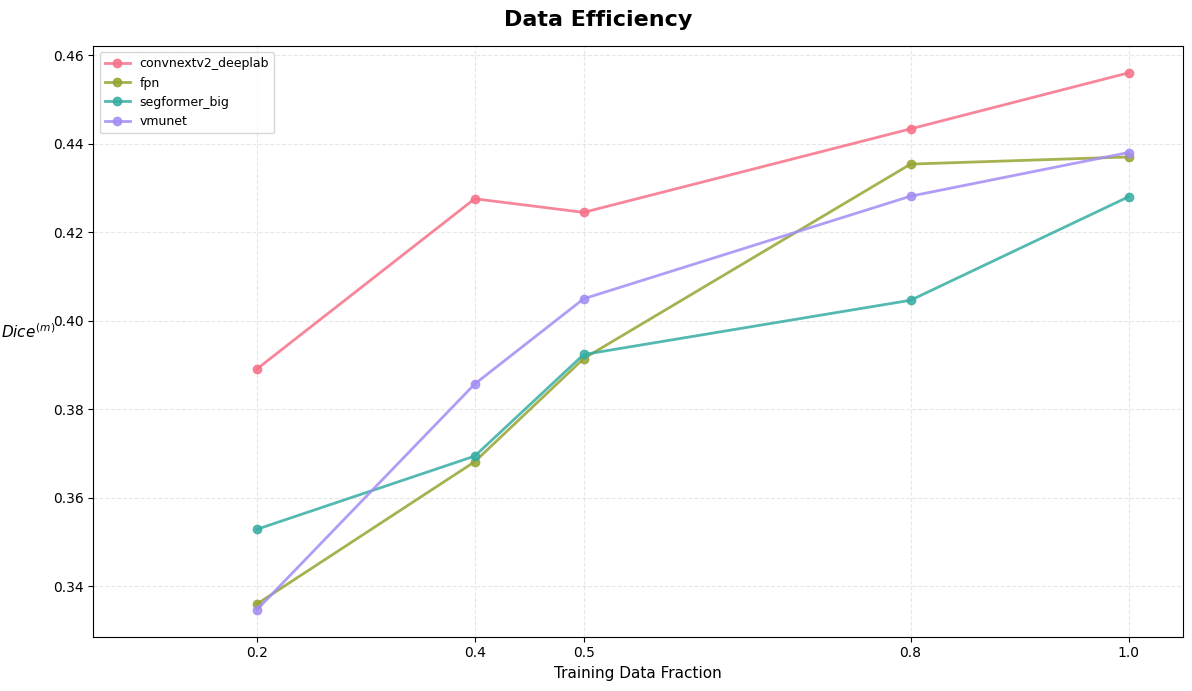}
\caption{Data efficiency scaling: Impact of training set volume on the predictive accuracy of top-performing architectures.}
\label{fig:dataefficiency}
\end{figure}

\begin{table*}[!th]
\scriptsize
\caption{Leave-One-Center-Out Cross-Validation results assessing out-of-distribution generalization. $\pm$ denotes the 95\% bootstrap confidence interval.}
\label{tab:leave_one_center_out_table}
\setlength{\tabcolsep}{3pt}
\centering
\begin{tabular}{|c|c|c|c|c|c|c|}
\hline
\multirow{2}{*}{\makecell{\textbf{Hold-out} \\ \textbf{Clinical Center}}} & \multirow{2}{*}{\makecell{\textbf{Training} \\ \textbf{Set Size}}} & \multirow{2}{*}{\makecell{\textbf{Testing} \\ \textbf{Set Size}}} & \multicolumn{4}{c|}{\textbf{$\mathit{Dice}^{(m)}$}} \\ \cline{4-7}
 & & & \textbf{ConvNeXt V2 DeepLab} & \textbf{FPN} & \textbf{SegFormer (Big)} & \textbf{VMUnet} \\
\hline
Center 1 & 606 & 39 & $\mathbf{0.539}_{{\pm 0.047}}$ & $\mathbf{0.541}_{{\pm 0.054}}$ & $\mathbf{0.466}_{{\pm 0.043}}$ & $0.480_{{\pm 0.048}}$ \\
Center 2 & 279 & 366 & $\mathbf{0.380}_{{\pm 0.016}}$ & $\mathbf{0.398}_{{\pm 0.017}}$ & $\mathbf{0.367}_{{\pm 0.016}}$ & $\mathbf{0.387}_{{\pm 0.015}}$ \\
Center 3 & 497 & 148 & $0.472_{{\pm 0.023}}$ & $0.425_{{\pm 0.024}}$ & $0.428_{{\pm 0.023}}$ & $0.448_{{\pm 0.024}}$ \\
Center 4 & 564 & 81 & $0.458_{{\pm 0.030}}$ & $0.400_{{\pm 0.032}}$ & $0.406_{{\pm 0.030}}$ & $0.390_{{\pm 0.032}}$ \\
Center 5 & 634 & 11 & $0.488_{{\pm 0.083}}$ & $0.510_{{\pm 0.059}}$ & $0.459_{{\pm 0.090}}$ & $\mathbf{0.483}_{{\pm 0.090}}$ \\ \hline
\multicolumn{3}{|c|}{\textbf{Average}} & 0.467 & 0.455 & 0.425 & 0.438 \\ \hline
\end{tabular}

\end{table*}

\textbf{Computational Efficiency:}
As illustrated in \cref{fig:flops_f1}, vertical parameter scaling does not strictly correlate with improved segmentation accuracy. The heaviest architectures (\eg, ConvNeXt + UPerNet) yield sub-par metrics compared to lighter models. Mamba-UNet emerges as exceptionally efficient, residing at the lower end of the GFLOPs spectrum while achieving state-of-the-art diameter preservation. \\

\textbf{Qualitative Analysis:}
The quantitative metrics are reflected in \cref{fig:qualitative_comparison}, which depicts predictions across selected architectural families. The visualizations show characteristic failure modes on complex topologies. RepLKNet paired with DeepLab exhibits severe over-dilation, failing to resolve the fine-grained boundaries of distal branches. Conversely, VM-UNet (SSM) demonstrates excellent morphological fidelity, though it occasionally fragments on the thinnest distal tips. ConvNeXt V2 preserves spatial contiguity better across branching intersections. Ultimately, the ensemble yields the cleanest segmentations, effectively suppressing the false-positive background noise seen in individual models while maintaining unbroken topological centerlines along the most challenging peripheral arteries.

\subsection{Cross-Center Generalization}
\Cref{tab:leave_one_center_out_table} presents the Leave-One-Center-Out generalization experiments. While absolute performance appears to fluctuate across centers (\eg, a visible drop when Center 2 is isolated), this variation is confounded with the training set size, which differs across held-out centers; we therefore cannot attribute it to domain or data quantity alone. Furthermore, center 5 contains only 11 frames, so its quantitative estimate is of limited reliability. Results on the remaining centers are broadly in line with the baseline averages, which provides modest evidence of out-of-distribution robustness.

\subsection{Data Efficiency}

The data scaling experiments (\cref{fig:dataefficiency}) demonstrate that dataset size remains a bottleneck for this task. The ConvNeXt V2 + DeepLabV3+ model reaches the 100\%-data performance of most other architectures using only 50\% of the training volume, but its own performance continues to improve with more data. The sustained upward trend across all models indicates that none of them has saturated. Future improvements in this domain will likely be driven as much by expanding dataset scale as by novel architectural design.

\section{Conclusion, Limitations \& Future Work}
CARDIAG provides a rigorously annotated, leakage-free, multi-center benchmark for dense SYNTAX segmentation, overcoming the field's reliance on irreproducible private data and public splits that cannot be checked for patient leakage. Our 24-architecture evaluation establishes robust baselines, revealing that modern CNNs (ConvNeXt V2) achieve among the best mask overlap ($Dice$), while the State-Space Model Mamba-UNet obtains the lowest diameter error.

\textbf{Limitations:} Current baselines process frames independently, failing to exploit the dataset's temporal sequences. The benchmark itself also has several limitations. Firstly, the uncertainty masks are used neither in training nor in the benchmark metrics, so we do not separate annotator uncertainty from model error. Secondly, annotation reliability is reported only as Dice on ten cases for the entire binary masks; per-SYNTAX-class inter-annotator agreement is not reported. Additionally, mitigating observed demographic and projection-based biases is left to future work. Finally, while our 645-frame scale is more compact than coarser datasets like ARCADE, it deliberately trades raw volume for, to our knowledge, unprecedented, expert-driven annotation density. Our data efficiency analysis shows that performance still improves with additional training data and has not saturated, indicating that larger datasets would further benefit this task, even though the current scale already supports a meaningful architectural comparison.

\textbf{Community Impact \& Clinical Relevance:} Releasing CARDIAG pushes the community beyond binary vessel extraction toward fine-grained segmentation and diameter preservation. This high-fidelity data will accelerate the development of automated quantitative coronary angiography (QCA) systems and end-to-end SYNTAX score calculators, ultimately facilitating clinical decision-making for PCI versus CABG interventions.

\subsection*{Acknowledgments}
The research was supported in part by project “Cloud Artificial Intelligence Service Engineering
(CAISE) platform to create universal and smart services for various application areas”, No.
KPOD.05.10-IW.10-0005/24, as part of the European IPCEI-CIS program, financed by NRRP
(National Recovery and Resilience Plan) funds.

{
    \small
    \bibliographystyle{ieeenat_fullname}
    \bibliography{main}

@String(CVPR= {IEEE Conf. Comput. Vis. Pattern Recog.})

@String(ICCV= {Int. Conf. Comput. Vis.})

@String(ECCV= {Eur. Conf. Comput. Vis.})

@String(CVPR  = {CVPR})

@String(ICCV  = {ICCV})

@String(ECCV  = {ECCV})

@article{Kaba2023TheAO,
  title = {The Application of Deep Learning for the Segmentation and Classification of Coronary Arteries},
  volume = {13},
  number = {13},
  journal = {Diagnostics},
  author = {Kaba, Serife and Haci, Huseyin and Isin, Ali and Ilhan, Ahmet and Conkbayir, Cenk},
  year = {2023},
  month = jul,
  pages = {2274},
  note = {doi: 10.3390/diagnostics13132274}
}

@article{ARCADE,
  title = {Dataset for Automatic Region-based Coronary Artery Disease Diagnostics Using {X-Ray} Angiography Images},
  volume = {11},
  number = {1},
  journal = {Sci. Data},
  author = {Popov, Maxim and others},
  year = {2024},
  month = jan,
  note = {doi: 10.1038/s41597-023-02871-z}
}

@article{PPL,
  title = {Progressive Perception Learning for Main Coronary Segmentation in {X-Ray} Angiography},
  volume = {42},
  number = {3},
  journal = {IEEE Trans. Med. Imaging},
  author = {Zhang, Hongwei and Gao, Zhifan and Zhang, Dong and Hau, William Kongto and Zhang, Heye},
  year = {2023},
  month = mar,
  pages = {864--879},
  note = {doi: 10.1109/TMI.2022.3219126}
}

@article{HAGMNUQ,
  title = {{HAGMN-UQ}: Hyper association graph matching network with uncertainty quantification for coronary artery semantic labeling},
  volume = {99},
  journal = {Med. Image Anal.},
  author = {Zhao, Chen and Esposito, Michele and Xu, Zhihui and Zhou, Weihua},
  year = {2025},
  month = jan,
  pages = {103374},
  note = {doi: 10.1016/j.media.2024.103374}
}

@article{AGMN,
  title = {{AGMN}: Association graph-based graph matching network for coronary artery semantic labeling on invasive coronary angiograms},
  volume = {143},
  journal = {Pattern Recognit.},
  author = {Zhao, Chen and others},
  year = {2023},
  month = nov,
  pages = {109789},
  note = {doi: 10.1016/j.patcog.2023.109789}
}

@article{EAGMN,
  title = {{EAGMN}: Coronary artery semantic labeling using edge attention graph matching network},
  volume = {166},
  journal = {Comput. Biol. Med.},
  author = {Zhao, Chen and Xu, Zhihui and Hung, Guang-Uei and Zhou, Weihua},
  year = {2023},
  month = nov,
  pages = {107469},
  note = {doi: 10.1016/j.compbiomed.2023.107469}
}

@article{MGM,
  title = {Multi-graph graph matching for coronary artery semantic labeling in invasive coronary angiograms},
  volume = {169},
  journal = {Pattern Recognit.},
  author = {Zhao, Chen and Xu, Zhihui and Baral, Pukar and Esposito, Michel and Zhou, Weihua},
  year = {2026},
  month = jan,
  pages = {111943},
  note = {doi: 10.1016/j.patcog.2025.111943}
}

@article{Park2023,
  title = {Selective ensemble methods for deep learning segmentation of major vessels in invasive coronary angiography},
  volume = {50},
  number = {12},
  journal = {Med. Phys.},
  author = {Park, Jeeone and others},
  year = {2023},
  month = jun,
  pages = {7822--7839},
  note = {doi: 10.1002/mp.16554}
}

@article{Xian2020,
  title = {Main Coronary Vessel Segmentation Using Deep Learning in Smart Medical},
  volume = {2020},
  journal = {Math. Probl. Eng.},
  author = {Xian, Zhanchao and Wang, Xiaoqing and Yan, Shaodi and Yang, Dahao and Chen, Junyu and Peng, Changnong},
  year = {2020},
  month = oct,
  pages = {1--9},
  note = {doi: 10.1155/2020/8858344}
}

@article{Zhang2026,
  title = {{HR-UMamba++}: A High-Resolution Multi-Directional Mamba Framework for Coronary Artery Segmentation in {X-Ray} Coronary Angiography},
  volume = {10},
  number = {1},
  journal = {Fractal Fract.},
  author = {Zhang, Xiuhan and Lu, Peng and Zheng, Zongsheng and Li, Wenhui},
  year = {2026},
  month = jan,
  pages = {43},
  note = {doi: 10.3390/fractalfract10010043}
}

@article{Jun2020,
  title = {{T-Net}: Nested encoder--decoder architecture for the main vessel segmentation in coronary angiography},
  volume = {128},
  journal = {Neural Netw.},
  author = {Jun, Tae Joon and Kweon, Jihoon and Kim, Young-Hak and Kim, Daeyoung},
  year = {2020},
  month = aug,
  pages = {216--233},
  note = {doi: 10.1016/j.neunet.2020.05.002}
}

@article{radimagenet,
  title = {{RadImageNet}: An Open Radiologic Deep Learning Research Dataset for Effective Transfer Learning},
  volume = {4},
  number = {5},
  journal = {Radiol. Artif. Intell.},
  author = {Mei, Xueyan and others},
  year = {2022},
  month = sep,
  note = {doi: 10.1148/ryai.210315}
}

@article{SYNTAXtax,
  title = {The {SYNTAX} Score: An angiographic tool grading the complexity of coronary artery disease},
  volume = {1},
  number = {2},
  journal = {EuroIntervention},
  author = {Sianos, Georgios and others},
  year = {2005},
  month = aug,
  pages = {219--227}
}

@article{Yang2026,
  title = {Accurate segmentation and labeling of coronary artery segments in {X-Ray} angiography with an improved {UNet}-based {cGAN} architecture},
  volume = {112},
  journal = {Biomed. Signal Process. Control},
  author = {Yang, Qiuju and Yi, Hang and Yi, Liangping and Liu, Mian and Chen, Xuliang},
  year = {2026},
  month = feb,
  pages = {108812},
  note = {doi: 10.1016/j.bspc.2025.108812}
}

@article{Zhu2021,
  title = {Coronary angiography image segmentation based on {PSPNet}},
  volume = {200},
  journal = {Comput. Methods Programs Biomed.},
  author = {Zhu, Xiliang and Cheng, Zhaoyun and Wang, Sheng and Chen, Xianjie and Lu, Guoqing},
  year = {2021},
  month = mar,
  pages = {105897},
  note = {doi: 10.1016/j.cmpb.2020.105897}
}

@article{Xu2024,
  title = {{G2ViT}: Graph Neural Network-Guided Vision Transformer Enhanced Network for retinal vessel and coronary angiograph segmentation},
  volume = {176},
  journal = {Neural Netw.},
  author = {Xu, Hao and Wu, Yun},
  year = {2024},
  month = aug,
  pages = {106356},
  note = {doi: 10.1016/j.neunet.2024.106356}
}

@article{Chang2024,
  title = {Optimizing ensemble {U-Net} architectures for robust coronary vessel segmentation in angiographic images},
  volume = {14},
  number = {1},
  journal = {Sci. Rep.},
  author = {Chang, Shih-Sheng and others},
  year = {2024},
  month = mar,
  note = {doi: 10.1038/s41598-024-57198-5}
}

@article{Du2021,
  title = {Training and validation of a deep learning architecture for the automatic analysis of coronary angiography},
  volume = {17},
  number = {1},
  journal = {EuroIntervention},
  author = {Du, Tianming and others},
  year = {2021},
  month = may,
  pages = {32--40},
  note = {doi: 10.4244/eij-d-20-00570}
}

@article{CervantesSanchez2019,
  title = {Automatic Segmentation of Coronary Arteries in {X-Ray} Angiograms using Multiscale Analysis and Artificial Neural Networks},
  volume = {9},
  number = {24},
  journal = {Appl. Sci.},
  author = {Cervantes-Sanchez, Fernando and others},
  year = {2019},
  month = dec,
  pages = {5507},
  note = {doi: 10.3390/app9245507}
}

@article{Kruzhilov2025,
  title = {{CoronaryDominance}: Angiogram dataset for coronary dominance classification},
  volume = {12},
  number = {1},
  journal = {Sci. Data},
  author = {Kruzhilov, Ivan and others},
  year = {2025},
  month = feb,
  note = {doi: 10.1038/s41597-025-04676-8}
}

@article{Palaniappan2026,
  title = {2026 Heart Disease and Stroke Statistics: A Report of {US} and Global Data From the {American Heart Association}},
  volume = {153},
  number = {9},
  journal = {Circulation},
  author = {Palaniappan, Latha P. and others},
  year = {2026},
  month = mar,
  note = {doi: 10.1161/cir.0000000000001412}
}

@article{Dodge1992,
  title = {Lumen diameter of normal human coronary arteries. Influence of age, sex, anatomic variation, and left ventricular hypertrophy or dilation.},
  volume = {86},
  number = {1},
  journal = {Circulation},
  author = {Dodge, J T and Brown, B G and Bolson, E L and Dodge, H T},
  year = {1992},
  month = jul,
  pages = {232--246},
  note = {doi: 10.1161/01.cir.86.1.232}
}

@inproceedings{xrayvision,
  title = {{TorchXRayVision}: A library of chest {X-Ray} datasets and models},
  author = {Cohen, Joseph Paul and others},
  booktitle = {Proc. Med. Image Deep Learn. (MIDL)},
  address = {Zurich, Switzerland},
  year = {2022}
}

@inproceedings{FPN,
  title = {Feature Pyramid Networks for Object Detection},
  author = {Lin, Tsung-Yi and Dollar, Piotr and Girshick, Ross and He, Kaiming and Hariharan, Bharath and Belongie, Serge},
  booktitle = {Proc. IEEE Conf. Comput. Vis. Pattern Recognit. (CVPR)},
  address = {Honolulu, HI, USA},
  year = {2017},
  month = jul,
  pages = {936--944},
  note = {doi: 10.1109/CVPR.2017.106}
}

@inproceedings{PSPNet,
  title = {Pyramid Scene Parsing Network},
  author = {Zhao, Hengshuang and Shi, Jianping and Qi, Xiaojuan and Wang, Xiaogang and Jia, Jiaya},
  booktitle = {Proc. IEEE Conf. Comput. Vis. Pattern Recognit. (CVPR)},
  address = {Honolulu, HI, USA},
  year = {2017},
  month = jul,
  pages = {6230--6239},
  note = {doi: 10.1109/CVPR.2017.660}
}

@inproceedings{DEEPLABV3PLUS,
  title = {Encoder-Decoder with Atrous Separable Convolution for Semantic Image Segmentation},
  author = {Chen, Liang-Chieh and Zhu, Yukun and Papandreou, George and Schroff, Florian and Adam, Hartwig},
  booktitle = {Proc. Eur. Conf. Comput. Vis. (ECCV)},
  address = {Munich, Germany},
  year = {2018},
  pages = {833--851},
  note = {doi: 10.1007/978-3-030-01234-2\_49}
}

@inproceedings{UPERNET,
  title = {Unified Perceptual Parsing for Scene Understanding},
  author = {Xiao, Tete and Liu, Yingcheng and Zhou, Bolei and Jiang, Yuning and Sun, Jian},
  booktitle = {Proc. Eur. Conf. Comput. Vis. (ECCV)},
  address = {Munich, Germany},
  year = {2018},
  pages = {432--448},
  note = {doi: 10.1007/978-3-030-01228-1\_26}
}

@inproceedings{oktay2018attention,
  title = {{Attention U-Net}: Learning Where to Look for the Pancreas},
  author = {Oktay, Ozan and others},
  booktitle = {Proc. Med. Image Deep Learn. (MIDL)},
  address = {Amsterdam, The Netherlands},
  year = {2018}
}

@inproceedings{unet,
  title = {{U-Net}: Convolutional Networks for Biomedical Image Segmentation},
  author = {Ronneberger, Olaf and Fischer, Philipp and Brox, Thomas},
  booktitle = {Proc. Med. Image Comput. Comput.-Assist. Interv. (MICCAI)},
  address = {Munich, Germany},
  year = {2015},
  pages = {234--241},
  note = {doi: 10.1007/978-3-319-24574-4\_28}
}

@inproceedings{segformer,
  title = {{SegFormer}: Simple and Efficient Design for Semantic Segmentation with Transformers},
  author = {Xie, Enze and Wang, Wenhai and Yu, Zhiding and Anandkumar, Anima and Alvarez, Jose M. and Luo, Ping},
  booktitle = {Adv. Neural Inform. Process. Syst. (NeurIPS)},
  year = {2021},
  pages = {12077--12090},
  adress = {online}
}

@inproceedings{swinunetr,
  title = {{SwinUNETR-V2}: Stronger Swin Transformers with Stagewise Convolutions for {3D} Medical Image Segmentation},
  author = {He, Yufan and Nath, Vishwesh and Yang, Dong and Tang, Yucheng and Myronenko, Andriy and Xu, Daguang},
  booktitle = {Proc. Med. Image Comput. Comput.-Assist. Interv. (MICCAI)},
  address = {Vancouver, BC, Canada},
  year = {2023},
  pages = {416--426},
  note = {doi: 10.1007/978-3-031-43901-8\_40}
}

@inproceedings{Shit2021,
  title = {{clDice} - a Novel Topology-Preserving Loss Function for Tubular Structure Segmentation},
  author = {Shit, Suprosanna and others},
  booktitle = {Proc. IEEE/CVF Conf. Comput. Vis. Pattern Recognit. (CVPR)},
  adress={online},
  year = {2021},
  month = jun,
  pages = {16555--16564},
  note = {doi: 10.1109/CVPR46437.2021.01629}
}

@inproceedings{convnext2,
  title = {{ConvNeXt V2}: Co-Designing and Scaling ConvNets With Masked Autoencoders},
  author = {Woo, Sanghyun and others},
  booktitle = {Proc. IEEE/CVF Conf. Comput. Vis. Pattern Recognit. (CVPR)},
  address = {Vancouver, BC, Canada},
  year = {2023},
  month = jun,
  pages = {16133--16142}
}

@inproceedings{segnext,
  title = {{SegNeXt}: Rethinking convolutional attention design for semantic segmentation},
  author = {Guo, Meng-Hao and Lu, Cheng-Ze and Hou, Qibin and Liu, Zheng-Ning and Cheng, Ming-Ming and Hu, Shi-Min},
  booktitle = {Adv. Neural Inform. Process. Syst. (NeurIPS)},
  address = {New Orleans, LA, USA},
  year = {2022}
}

@inproceedings{mednext,
  title = {{MedNeXt}: Transformer-Driven Scaling of ConvNets for Medical Image Segmentation},
  author = {Roy, Saikat and others},
  booktitle = {Proc. Med. Image Comput. Comput.-Assist. Interv. (MICCAI)},
  address = {Vancouver, BC, Canada},
  year = {2023},
  pages = {405--415},
  note = {doi: 10.1007/978-3-031-43901-8\_39}
}

@inproceedings{Izmailov2018AveragingWL,
  title = {Averaging Weights Leads to Wider Optima and Better Generalization},
  author = {Izmailov, Pavel and others},
  booktitle = {Proc. Conf. Uncertainty Artif. Intell. (UAI)},
  address = {Monterey, CA, USA},
  year = {2018}
}

@inproceedings{Shi2020,
  title = {{UENet}: A Novel Generative Adversarial Network for Angiography Image Segmentation},
  author = {Shi, Xiaotong and Du, Tianming and Chen, Shuang and Zhang, Honggang and Guan, Changdong and Xu, Bo},
  booktitle = {Proc. 42nd Annu. Int. Conf. IEEE Eng. Med. Biol. Soc. (EMBC)},
  address = {Montreal, QC, Canada},
  year = {2020},
  month = jul,
  pages = {1612--1615},
  note = {doi: 10.1109/EMBC44109.2020.9175334}
}

@inproceedings{Zhai2019,
  title = {Coronary Artery Vascular Segmentation on Limited Data via Pseudo-Precise Label},
  author = {Zhai, Mo and Du, Tianming and Yang, Ruolin and Zhang, Honggang},
  booktitle = {Proc. 41st Annu. Int. Conf. IEEE Eng. Med. Biol. Soc. (EMBC)},
  address = {Berlin, Germany},
  year = {2019},
  month = jul,
  pages = {816--819},
  note = {doi: 10.1109/EMBC.2019.8856682}
}

@inproceedings{CardioSyntax,
  title = {{CardioSyntax}: End-to-End {SYNTAX} Score Prediction - Dataset, Benchmark and Method},
  author = {Ponomarchuk, Alexander and others},
  booktitle = {Proc. IEEE/CVF Winter Conf. Appl. Comput. Vis. (WACV)},
  address = {Tucson, AZ, USA},
  year = {2025},
  month = feb,
  pages = {5873--5883},
  note = {doi: 10.1109/WACV61041.2025.00573}
}

@inproceedings{SeoSum_DiffusionBased_MICCAI2025,
  title = {Diffusion-Based User-Guided Data Augmentation for Coronary Stenosis Detection},
  author = {Seo, Sumin and Lee, In Kyu and Kim, Hyun-Woo and Min, Jaesik and Jung, Chung-Hwan},
  booktitle = {Proc. Med. Image Comput. Comput.-Assist. Interv. (MICCAI)},
  year = {2025},
  address= {Daejon, Republic of Korea},
  month = sep,
  pages = {149--169}
}

@inproceedings{Ma2021,
  title = {Self-Supervised Vessel Segmentation via Adversarial Learning},
  author = {Ma, Yuxin and others},
  booktitle = {Proc. IEEE/CVF Int. Conf. Comput. Vis. (ICCV)},
  address = {Montreal, QC, Canada},
  year = {2021},
  month = oct,
  pages = {7516--7525},
  note = {doi: 10.1109/ICCV48922.2021.00744}
}

@article{DEEPLABV3,
  title = {Rethinking Atrous Convolution for Semantic Image Segmentation},
  author = {Chen, Liang-Chieh and Papandreou, George and Schroff, Florian and Adam, Hartwig},
  journal = {arXiv preprint arXiv:1706.05587},
  year = {2017},
  note = {doi: 10.48550/ARXIV.1706.05587}
}

@article{replknet,
  title = {Scaling Up Your Kernels to 31x31: Revisiting Large Kernel Design in {CNNs}},
  author = {Ding, Xiaohan and Zhang, Xiangyu and Zhou, Yizhuang and Han, Jungong and Ding, Guiguang and Sun, Jian},
  journal = {arXiv preprint arXiv:2203.06717},
  year = {2022},
  note = {doi: 10.48550/ARXIV.2203.06717}
}

@article{ARCADE_Bilal,
  title = {Multivessel Coronary Artery Segmentation and Stenosis Localisation using Ensemble Learning},
  author = {Bilal, Muhammad and others},
  journal = {arXiv preprint arXiv:2310.17954},
  year = {2023},
  note = {doi: 10.48550/ARXIV.2310.17954}
}

@article{ARCADE_Ku,
  title = {{MPSeg} : Multi-Phase strategy for coronary artery Segmentation},
  author = {Ku, Jonghoe and Lee, Yong-Hee and Shin, Junsup and Lee, In Kyu and Kim, Hyun-Woo},
  journal = {arXiv preprint arXiv:2311.10306},
  year = {2023},
  note = {doi: 10.48550/ARXIV.2311.10306}
}

@article{ARCADE_Liu,
  title = {{YOLO-Angio}: An Algorithm for Coronary Anatomy Segmentation},
  author = {Liu, Tom and Lin, Hui and Katsaggelos, Aggelos K. and Kline, Adrienne},
  journal = {arXiv preprint arXiv:2310.15898},
  year = {2023},
  note = {doi: 10.48550/ARXIV.2310.15898}
}

@article{mambaunet,
  title = {{Mamba-UNet}: {UNet}-Like Pure Visual Mamba for Medical Image Segmentation},
  author = {Wang, Ziyang and Zheng, Jian-Qing and Zhang, Yichi and Cui, Ge and Li, Lei},
  journal = {arXiv preprint arXiv:2402.05079},
  year = {2024},
  note = {doi: 10.48550/ARXIV.2402.05079}
}

@article{VMUnet,
      title={{VM-UNet}: Vision Mamba UNet for Medical Image Segmentation}, 
      author={Jiacheng Ruan and Jincheng Li and Suncheng Xiang},
      journal = {arXiv preprint arXiv:2402.02491},
      year={2024},
      note = {doi: 10.48550/ARXIV.2402.02491}
}

@article{swinumamba,
      title={{Swin-UMamba}: Mamba-based UNet with ImageNet-based pretraining}, 
      author={Jiarun Liu and others},
      journal = {arXiv preprint arXiv:2402.03302},
      year={2024},
      note = {doi: 10.48550/ARXIV.2402.03302},
}

@article{geodl,
  title = {Geometric Deep Learning: Grids, Groups, Graphs, Geodesics, and Gauges},
  author = {Bronstein, Michael M. and Bruna, Joan and Cohen, Taco and Veli\v{c}kovi\'{c}, Petar},
  journal = {arXiv preprint arXiv:2104.13478},
  year = {2021},
  note = {doi: 10.48550/ARXIV.2104.13478}
}

@techreport{dicom,
  title = {Digital Imaging and Communications in Medicine ({DICOM})},
  author = {{National Electrical Manufacturers Association (NEMA)}},
  institution = {NEMA},
  address = {Rosslyn, VA, USA},
  year = {2026},
  note = {Standard PS3 / ISO 12052. [Online]. Available: https://dicom.nema.org/medical/dicom/2026b/}
}
}

\appendix

\renewcommand{\thetable}{S\arabic{table}}
\renewcommand{\thefigure}{S\arabic{figure}}
\renewcommand{\theHtable}{S\arabic{table}}   
\renewcommand{\theHfigure}{S\arabic{figure}}
\setcounter{table}{0}
\setcounter{figure}{0}

\section{Additional benchmark setup details}

\paragraph{Split composition}

\Cref{tab:split_center,tab:split_class} demonstrate the main benchmark split composition in terms of patient and center distribution and of class coverage.

\begin{table}[!th]
\centering
\footnotesize
\caption{Main benchmark split composition per center. Each cell gives the number of patients / the number of annotated primary frames. No patient appears in more than one split.}
\label{tab:split_center}
\begin{tabular}{@{}lrrrr@{}}
\hline
 & Train & Validation & Test & Total \\
\hline
Center 1 & 4 / 22 & 1 / 5 & 3 / 12 & 8 / 39 \\
Center 2 & 41 / 245 & 7 / 41 & 14 / 80 & 62 / 366 \\
Center 3 & 15 / 113 & 2 / 9 & 6 / 26 & 23 / 148 \\
Center 4 & 6 / 51 & 1 / 5 & 4 / 25 & 11 / 81 \\
Center 5 & 1 / 6 & 0 / 0 & 1 / 5 & 2 / 11 \\
\hline
\textbf{Total} & 67 / 437 & 11 / 60 & 28 / 148 & 106 / 645 \\
\hline
\end{tabular}
\end{table}

\begin{table}[!th]
\centering
\footnotesize
\caption{Class coverage of the main benchmark split: number of annotated primary frames containing each class, per split.}
\label{tab:split_class}
\begin{tabular}{@{}lrrrr@{}}
\hline
Class & Train & Validation & Test & Total \\
\hline
01 & 164 & 24 & 64 & 252 \\
02 & 163 & 25 & 64 & 252 \\
03 & 161 & 23 & 64 & 248 \\
04 & 148 & 23 & 57 & 228 \\
16 & 146 & 20 & 49 & 215 \\
16a & 97 & 15 & 44 & 156 \\
16b & 59 & 13 & 31 & 103 \\
16c & 22 & 4 & 12 & 38 \\
05 & 265 & 33 & 81 & 379 \\
06 & 269 & 35 & 82 & 386 \\
07 & 259 & 35 & 78 & 372 \\
08 & 233 & 33 & 76 & 342 \\
09 & 221 & 32 & 67 & 320 \\
09a & 86 & 23 & 25 & 134 \\
10 & 89 & 3 & 26 & 118 \\
10a & 17 & 2 & 2 & 21 \\
11 & 255 & 32 & 81 & 368 \\
12 & 92 & 4 & 26 & 122 \\
12a & 200 & 27 & 60 & 287 \\
13 & 204 & 28 & 62 & 294 \\
14 & 114 & 12 & 32 & 158 \\
15 & 12 & 0 & 4 & 16 \\
Catheter & 436 & 60 & 148 & 644 \\
\hline
\end{tabular}
\end{table}

Consequently, this main split is patient-disjoint, but not center-disjoint: every center appears in training, validation and test, so it measures generalization to unseen patients from known hospitals, not to unseen centers. Unseen-center generalization is also evaluated separately in the main text.

\paragraph{Model input}
In all benchmark experiments the network input is the single-channel primary frame, read as grayscale and scaled to $[0,1]$. Frames are used at their native $512\times512$ resolution, without resizing or cropping. The background acquisition, the temporal frames, the vesselness map and the acquisition metadata are not given to the models. In all runs, the checkpoint with the highest validation Dice score on the validation data is kept and evaluated.


\paragraph{Training augmentation}
During training we apply:
\begin{itemize}
    \item random rotation (up to $\pm0.3$ rad, probability 0.5)
    \item random elastic deformation (probability 0.2)
    \item random gamma contrast adjustment ($\gamma\in[0.7,1.5]$, probability 0.5)
    \item additive Gaussian noise (probability 0.2) 
    \item random zoom (scale $0.8$--$1.2$, probability 0.5)
\end{itemize}
Validation and test frames are not augmented. The batch size is 4.

\paragraph{Model targets}
The models predict 27 output channels: one background channel, 25 SYNTAX segment channels and one catheter channel. Three of the segment channels (12b, 14a, 14b) have no annotated examples in the dataset, so we report results only for the classes with non-zero cardinality, i.e., the 22 SYNTAX segments and the catheter that occur in the data.

\section{Dataset label statistics}

\begin{figure}
\centerline{\includegraphics[width=\columnwidth]{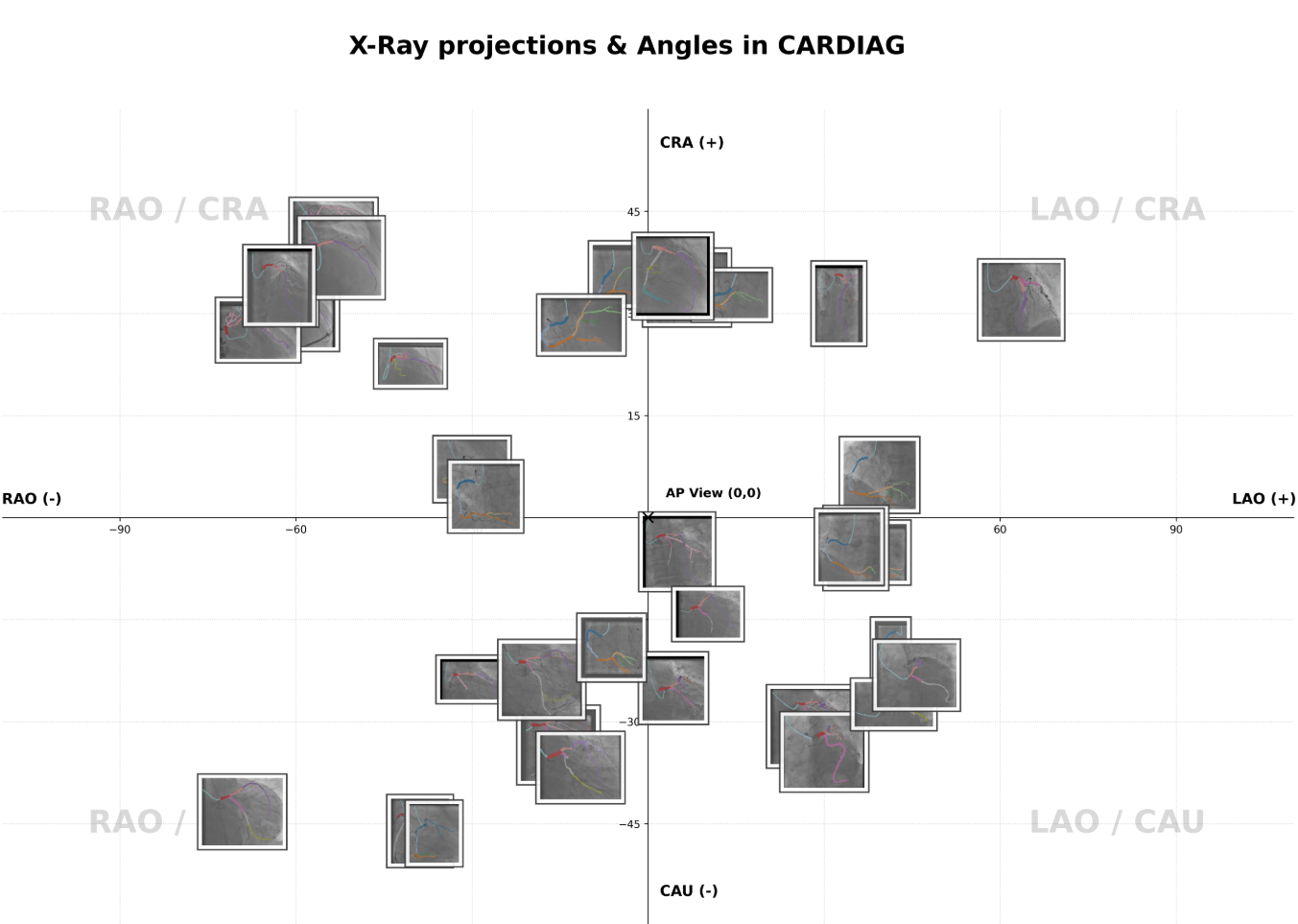}}
\caption{Labelled X-ray angiography images overlaid on ($\alpha$, $\beta$)-coordinate system.}
\label{fig:view_space}
\end{figure}

\begin{figure}
\centerline{\includegraphics[width=\columnwidth]{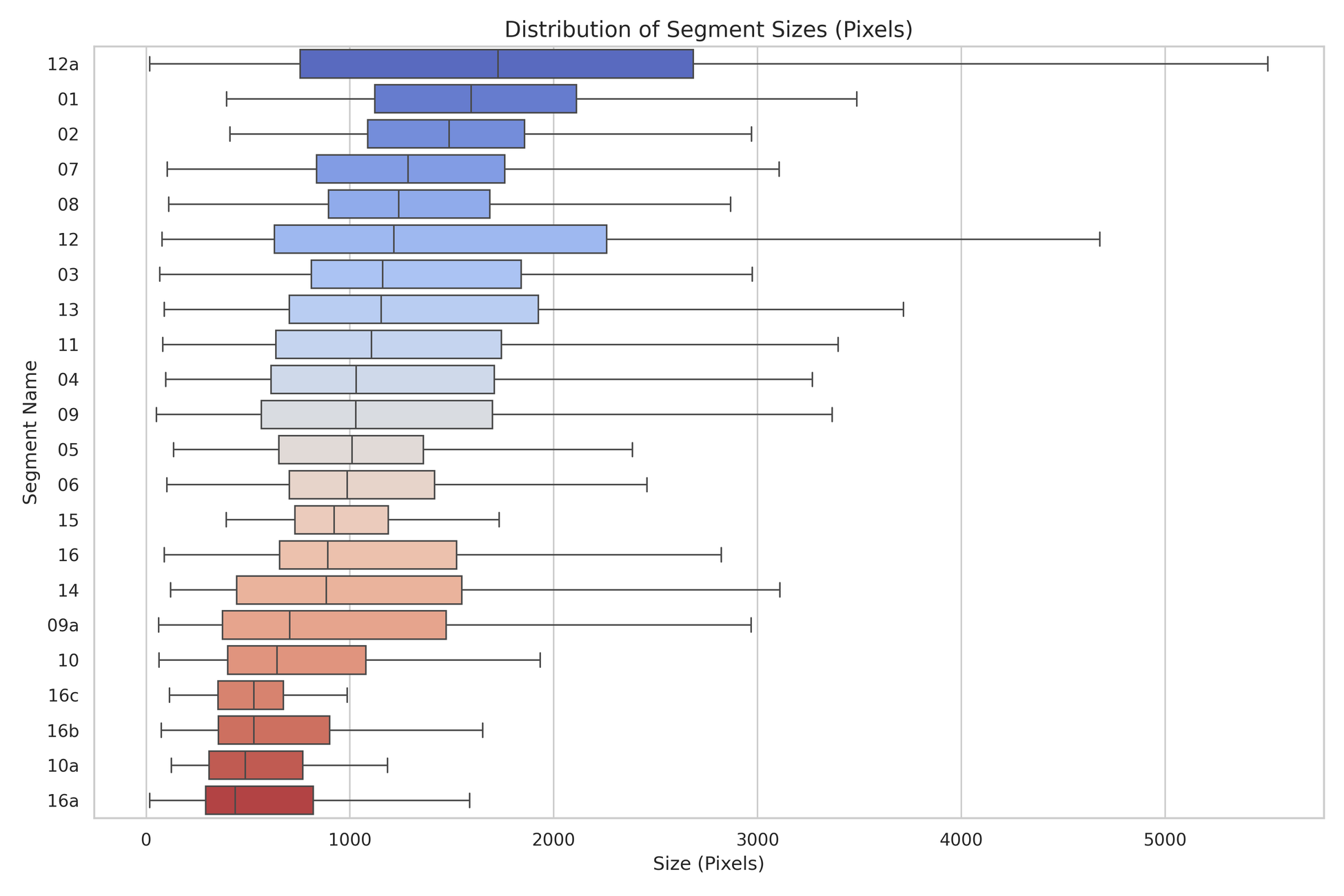}}
\caption{Average sizes of segments across dataset.}
\label{fig:segment_sizes}
\end{figure}

\begin{figure}
\centerline{\includegraphics[width=0.5\columnwidth]{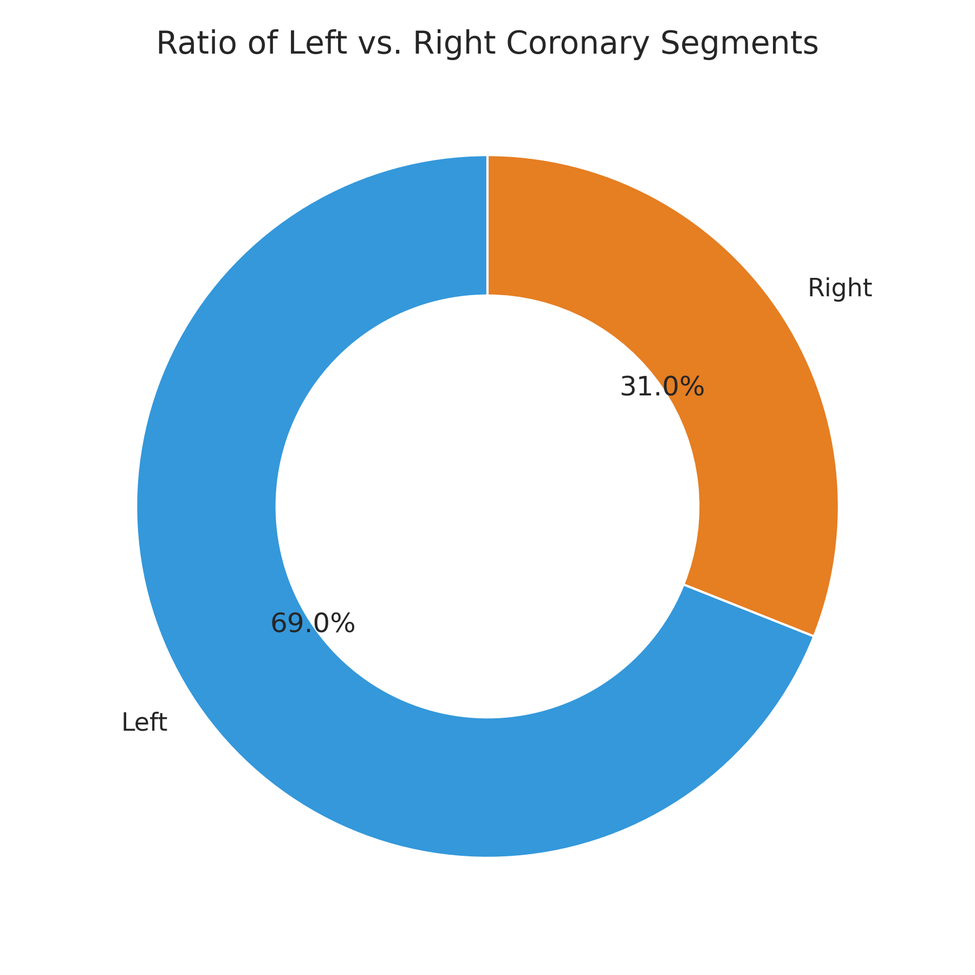}}
\caption{Left-to-right ratio of the provided data.}
\label{fig:lr_ratio}
\end{figure}

\begin{figure}
\centerline{\includegraphics[width=1.0\columnwidth]{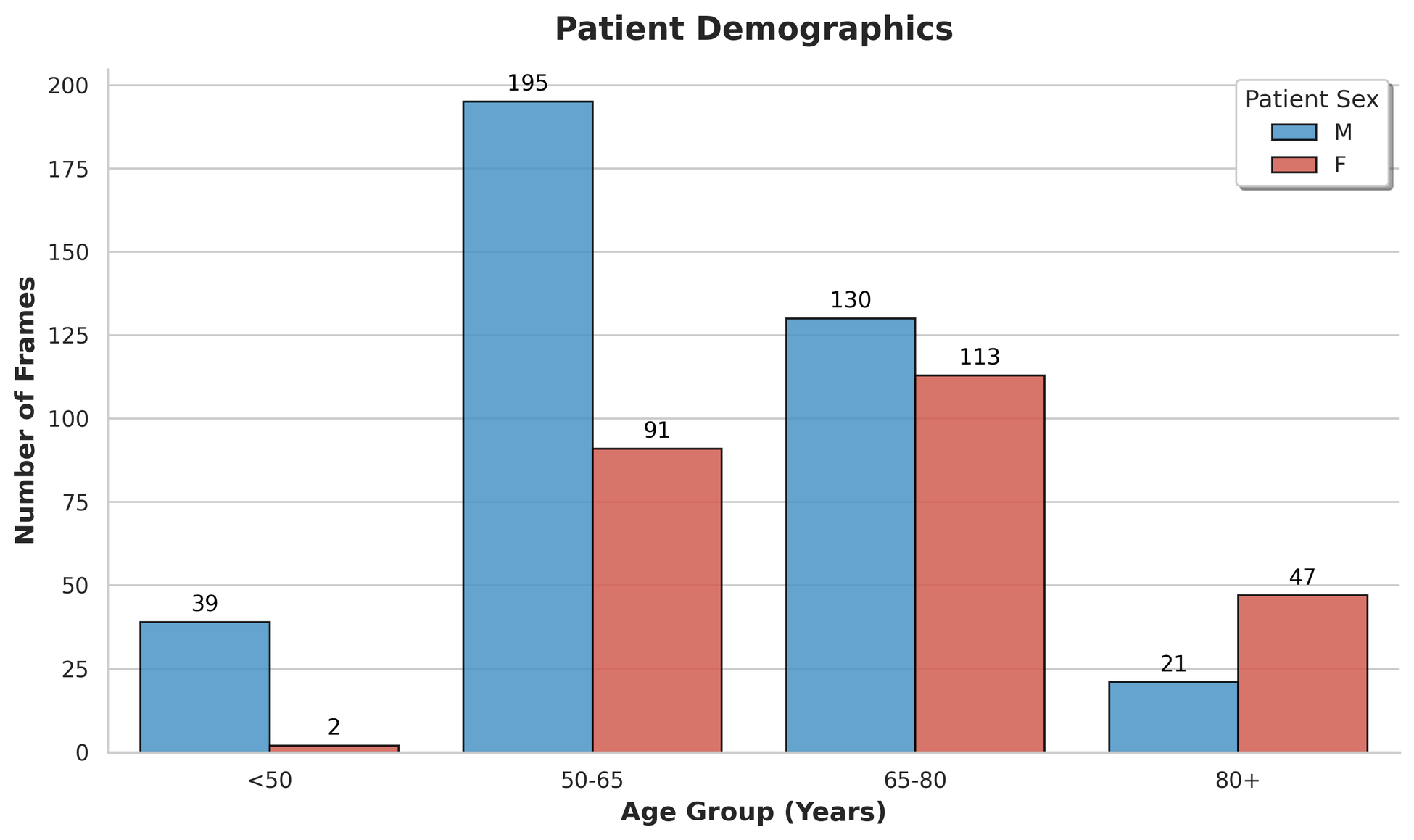}}
\caption{Patient demographics of the dataset: number of annotated frames per age group and sex.}
\label{fig:demography_stats}
\end{figure}

Label statistics are significant in reaching further conclusions in research based on our dataset. Firstly, the projection angles cover the most common settings of the C-arm as depicted in \cref{fig:view_space}. There is an imbalance in segment labels in a few domains: (1) as seen in \cref{fig:lr_ratio} there are more left sides than right sides - more videos are acquired for this particular side, it being harder to interpret (2) \cref{fig:segment_sizes} demonstrates that the more distal the segment, the smaller it typically is on the image. Furthermore, ultimately, some segments are not in the coronary trees of some patients, which is dependent on the heart dominance (\eg, segment 15 only appearing for people with left dominance). Additionally, the gathered data is not uniform in terms of age group magnitudes as mentioned, \eg, there are naturally more frames from patients in age groups where the coronary artery disease is most common (50-80 years). The entire human population is not reflected in the data and thus the models using the dataset as base should only be utilized for the aforementioned age range. This is visualized in \cref{fig:demography_stats}. Not only that, there are more men across the samples than women - statistically speaking men are more at risk of CAD and develop the condition earlier (thus the imbalance) \cite{Palaniappan2026}.

\section{Acquisition and Demographic Impacts}
\begin{figure}
\centerline{\includegraphics[width=\columnwidth]{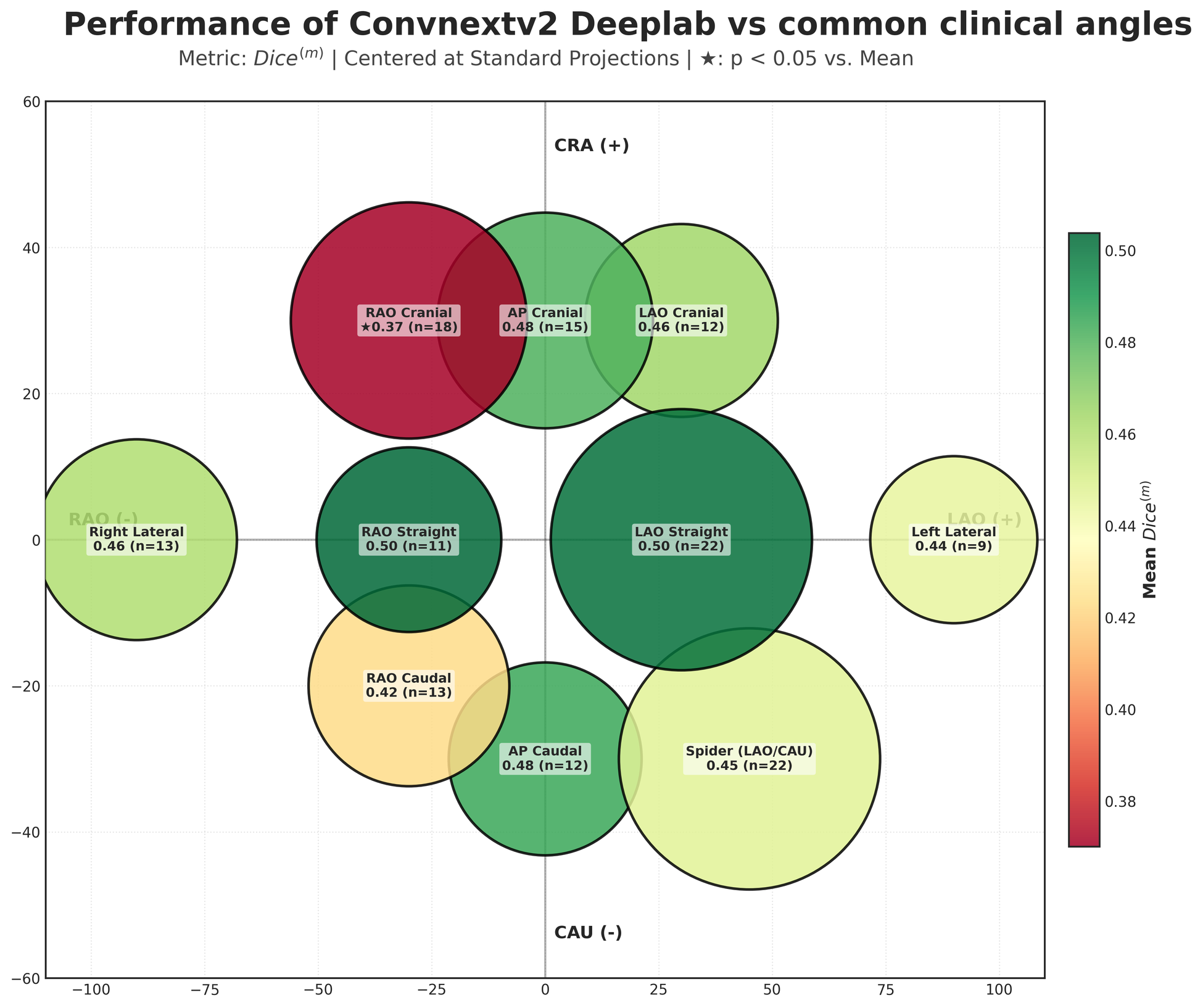}}
\caption{Dice performance mapped across C-arm acquisition perspectives ($\alpha, \beta$). Bubble size denotes sample frequency.}
\label{fig:angles_f1}
\end{figure}

\begin{figure}
\centerline{\includegraphics[width=\columnwidth]{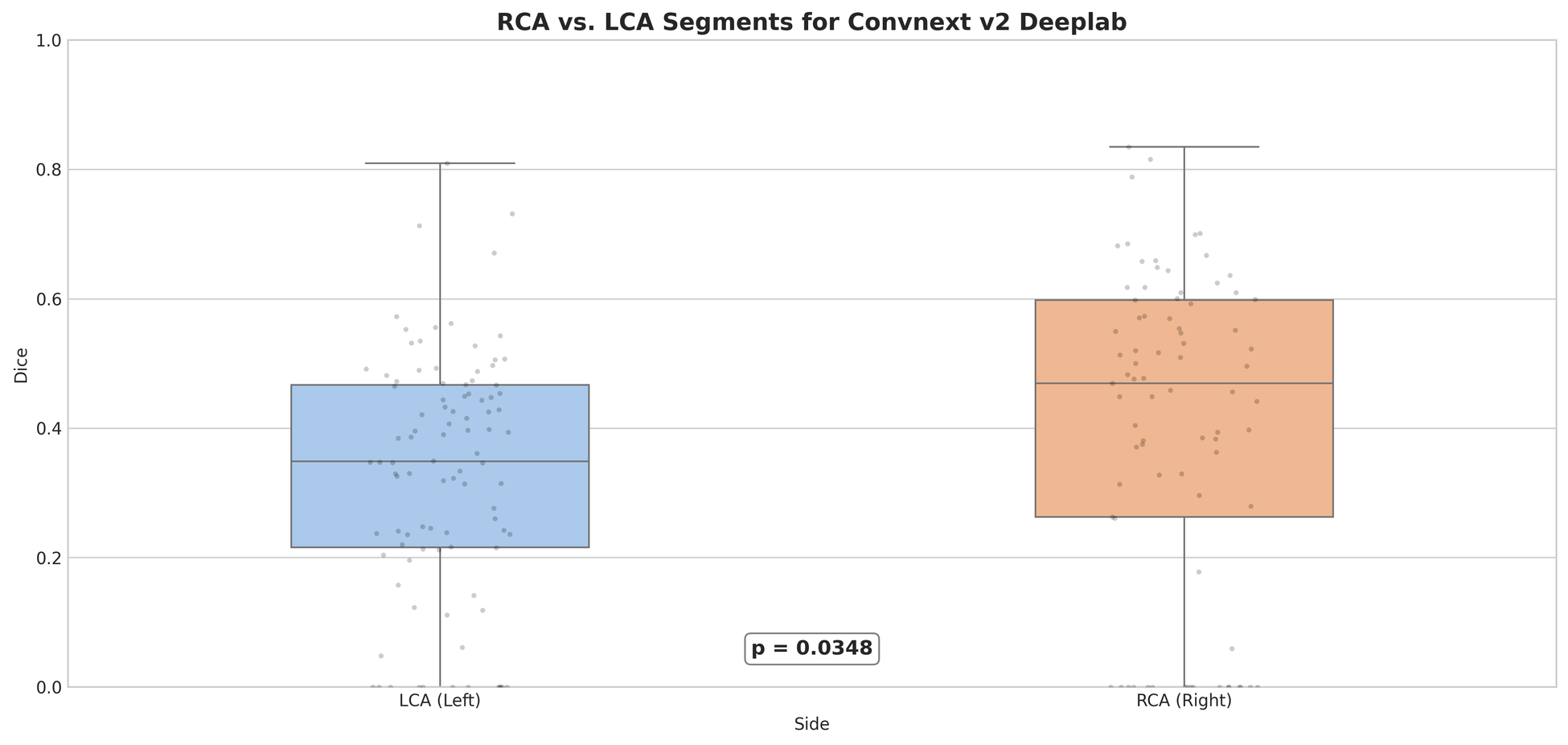}}
\caption{Dice score distribution comparing Left Coronary Artery (LCA) vs. Right Coronary Artery (RCA) segments.}
\label{fig:side_f1}
\end{figure}

\begin{figure}
\centerline{\includegraphics[width=\columnwidth]{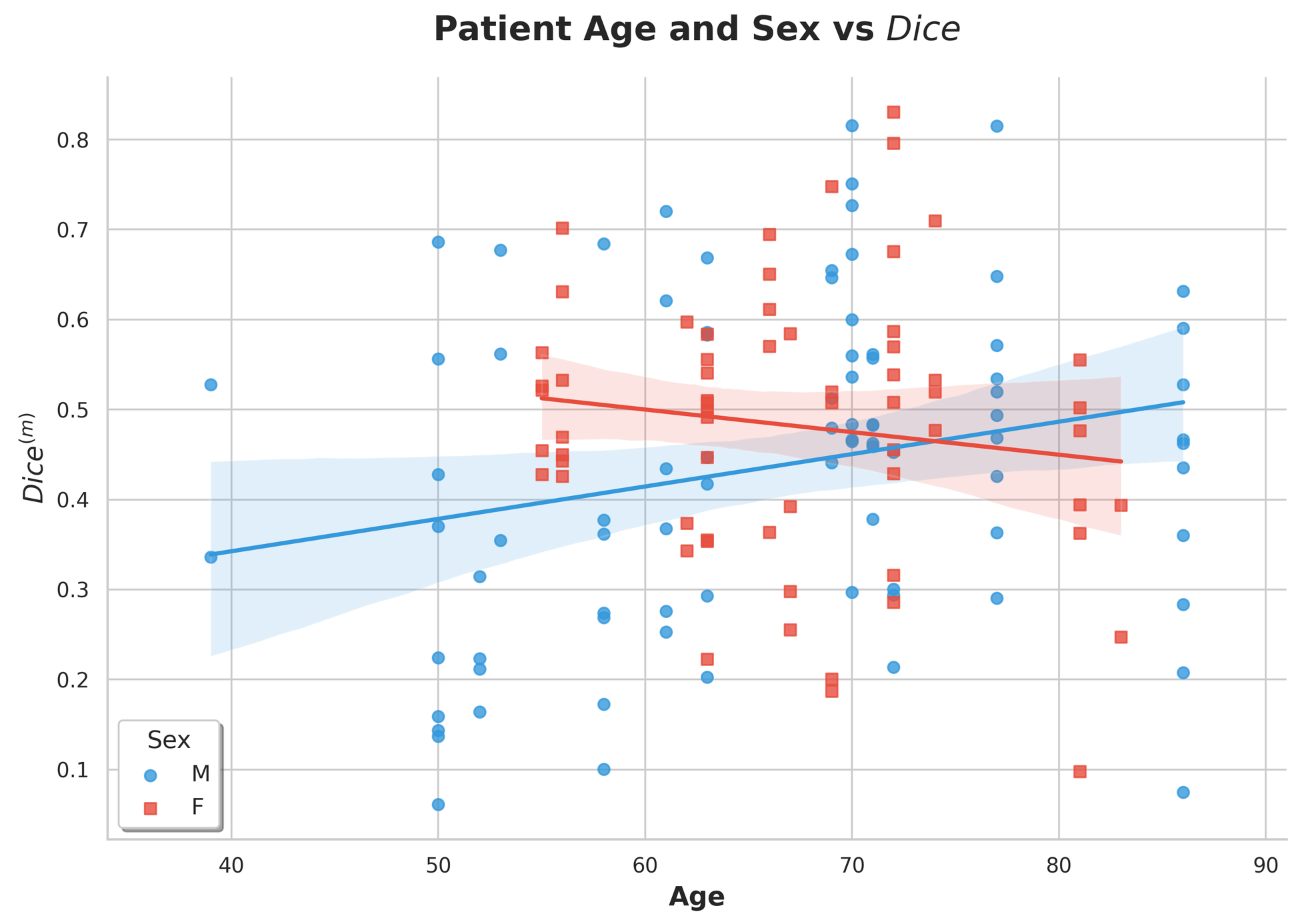}}
\caption{Algorithmic performance fluctuations correlated with patient demographics (Age and Sex).}
\label{fig:demographic_f1}
\end{figure}

Model performance exhibits distinct correlations with both clinical demographics and spatial acquisition parameters. 
First, the Left Coronary Artery (LCA) segments are segmented with lower accuracy than the Right (RCA) (\cref{fig:side_f1}), reflecting the higher topological complexity, branching density, and tortuosity of the left system. 
Second, mapping performance into the C-arm angular space (\cref{fig:angles_f1}) reveals that RAO Cranial views underperform the population average, primarily due to severe foreshortening and vessel overlap. Conversely, clear diagnostic views like LAO/RAO Straight overperform due to minimal anatomical distortion.
Finally, as shown in \cref{fig:demographic_f1}, models exhibit higher segmentation accuracy for male patients compared to females. This is likely driven by two factors: the higher statistical representation of male cases in the training distribution, and the established physiological baseline that male patients possess inherently larger coronary artery diameters \cite{Dodge1992}, which are visually easier to resolve.

\section{Additional benchmark results}

\begin{table*}[!th]
\scriptsize
\caption{Quantitative results of the ensembles of ConvNeXt V2, FPN and Mamba-UNet with different voting strategies. Results for all single models, and for the hard-voting ensemble, are in the main paper. Values are mean $\pm$ the 95\% bootstrap confidence interval.}
\label{tab:results_table}
\setlength{\tabcolsep}{3pt}
\begin{tabular*}{\textwidth}{@{\extracolsep{\fill}} l r r r r r r r r @{}}
\hline
Model & $\mathit{Dice}^{(m)}$ & $\mathit{Dice}^{(ci)}$ & $\mathit{Dice}^{(hi)}$ & $\mathit{Dice}^{(li)}$ & $DE$ & $clDice$ & $\mathcal{C}$ & $HD_{95}$ \\
\hline
Hard Voting & $0.479_{{\pm 0.026}}$ & $0.705_{{\pm 0.036}}$ & $0.493_{{\pm 0.045}}$ & $0.286_{{\pm 0.031}}$ & $2.428_{{\pm 0.130}}$ & $0.801_{{\pm 0.018}}$ & $0.981_{{\pm 0.003}}$ & $51.537_{{\pm 5.022}}$ \\
Soft Voting & $0.475_{{\pm 0.025}}$ & $0.708_{{\pm 0.035}}$ & $0.492_{{\pm 0.046}}$ & $0.285_{{\pm 0.031}}$ & $2.307_{{\pm 0.121}}$ & $0.815_{{\pm 0.017}}$ & $0.981_{{\pm 0.003}}$ & $52.032_{{\pm 5.033}}$ \\
Entropy Weighted & $0.468_{{\pm 0.025}}$ & $0.704_{{\pm 0.035}}$ & $0.483_{{\pm 0.045}}$ & $0.279_{{\pm 0.030}}$ & $2.296_{{\pm 0.120}}$ & $0.818_{{\pm 0.017}}$ & $0.981_{{\pm 0.003}}$ & $52.729_{{\pm 5.078}}$ \\
\hline
\end{tabular*}

\end{table*}

\begin{figure*}[p]
    \centering
    \includegraphics[
        width=1.0\textwidth,
        trim=20 10 20 10,
        clip
    ]{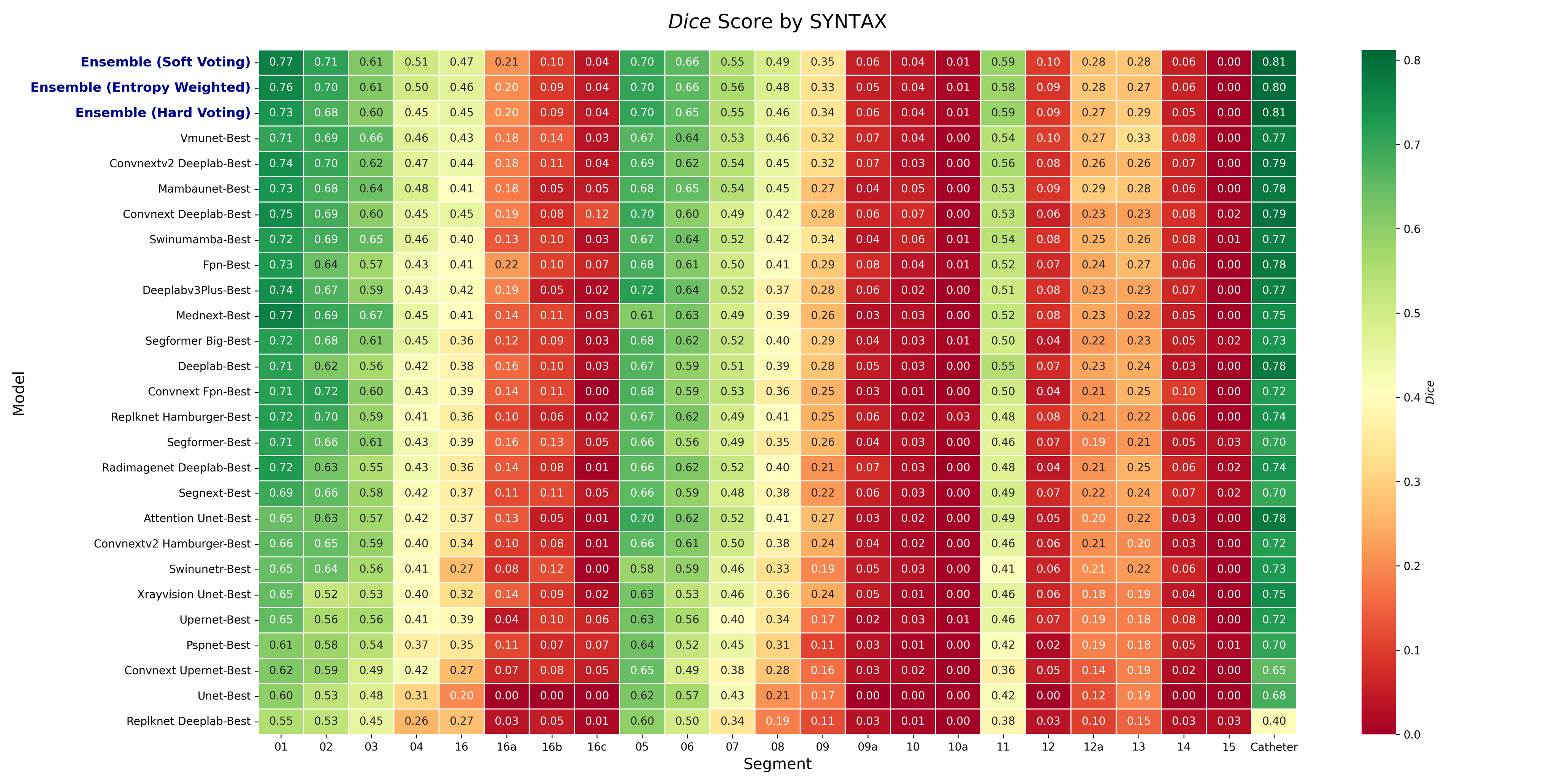}
    \caption{Architecture performance (Dice score) across individual SYNTAX segments, highlighting the steep performance drop on distal branches. Columns show the 23 output classes: 22 SYNTAX segments and the catheter.}
    \label{fig:persegment}
\end{figure*}

In \cref{tab:results_table}, we present the results of the ensembles comparison. Voting strategy has little effect on overlap: $Dice^{(m)}$ is 0.479, 0.475 and 0.468 for hard, soft and entropy-weighted voting, with overlapping confidence intervals. Soft and entropy-weighted voting, however, improve $clDice$ (0.815 and 0.818 vs. 0.801) and DE (2.307 and 2.296 vs. 2.428) over hard voting, and exceed the best single model on both (0.810 and 2.332). Hard voting is reported in the main paper because it achieves the highest $Dice^{(m)}$.

\Cref{fig:persegment} details performance across individual SYNTAX classes. While proximal segments (01, 02, 05) and the injection catheter are segmented with high fidelity across most architectures, there is a severe performance cliff for distal branches (\eg, 16a-c, 9a, 10, 14, 15), which approach zero Dice. As noted in Section~3 of the main paper, the distal parts of segments also exhibit the highest inter-observer variability among clinical experts. This performance cliff may be partly due to labeling ambiguity, but it is confounded with class scarcity (Figure~3 of the main paper) and model limitations.

\section{Statistical Significance Analysis}

To assess whether the performance differences between the best-performing architectures and the remaining models are statistically significant, we applied the Wilcoxon signed-rank test on paired per-image $\mathit{Dice}^{(m)}$ scores, comparing the best ensemble (Hard Voting) and the best single model (ConvNeXt V2 DeepLab) against every other architecture in the benchmark ($N=24$). Since this constitutes a family of multiple pairwise comparisons per reference model, we applied Holm's correction to control the family-wise error rate. \Cref{tab:statistical_significance} reports the Holm-corrected $p$-values and the resulting significance ($p_{\mathrm{Holm}} < 0.05$) for both reference models. All comparisons against both the best ensemble and the best single model were statistically significant.

\begin{table*}[!th]
\scriptsize
\caption{Wilcoxon signed-rank test results (Holm-corrected) for $\mathit{Dice}^{(m)}$. We compare both the best-performing ensemble (Hard Voting) and the best single model (ConvNeXt V2 DeepLab) against every other model in the benchmark.}
\label{tab:statistical_significance}
\setlength{\tabcolsep}{3pt}
\begin{tabular*}{\textwidth}{@{\extracolsep{\fill}} l rr rr @{}}
\hline
 & \multicolumn{2}{c}{vs.\ Best model} & \multicolumn{2}{c}{vs.\ Ensemble} \\
\cline{2-3} \cline{4-5}
Model & $p_{\mathrm{Holm}}$ & Sig. & $p_{\mathrm{Holm}}$ & Sig. \\
\hline
ConvNeXt V2 DeepLab & -- & -- & $\mathbf{<0.001}$ & Yes \\
VMUnet & $\mathbf{0.025}$ & Yes & $\mathbf{<0.001}$ & Yes \\
FPN & $\mathbf{0.025}$ & Yes & $\mathbf{<0.001}$ & Yes \\
DeepLab V3+ & $\mathbf{0.012}$ & Yes & $\mathbf{<0.001}$ & Yes \\
SegFormer (Big) & $\mathbf{<0.001}$ & Yes & $\mathbf{<0.001}$ & Yes \\
Swin-UMamba & $\mathbf{<0.001}$ & Yes & $\mathbf{<0.001}$ & Yes \\
Mamba-UNet & $\mathbf{0.001}$ & Yes & $\mathbf{<0.001}$ & Yes \\
RepLKNet Hamburger & $\mathbf{<0.001}$ & Yes & $\mathbf{<0.001}$ & Yes \\
ConvNeXt FPN & $\mathbf{<0.001}$ & Yes & $\mathbf{<0.001}$ & Yes \\
ConvNeXt DeepLab & $\mathbf{<0.001}$ & Yes & $\mathbf{<0.001}$ & Yes \\
DeepLab & $\mathbf{<0.001}$ & Yes & $\mathbf{<0.001}$ & Yes \\
Rad ImageNet DeepLab & $\mathbf{<0.001}$ & Yes & $\mathbf{<0.001}$ & Yes \\
MedNeXt & $\mathbf{<0.001}$ & Yes & $\mathbf{<0.001}$ & Yes \\
Segformer & $\mathbf{<0.001}$ & Yes & $\mathbf{<0.001}$ & Yes \\
ConvNeXt V2 Hamburger & $\mathbf{<0.001}$ & Yes & $\mathbf{<0.001}$ & Yes \\
Attention UNet & $\mathbf{<0.001}$ & Yes & $\mathbf{<0.001}$ & Yes \\
SegNeXt & $\mathbf{<0.001}$ & Yes & $\mathbf{<0.001}$ & Yes \\
Swin-UNETR & $\mathbf{<0.001}$ & Yes & $\mathbf{<0.001}$ & Yes \\
PSPNet & $\mathbf{<0.001}$ & Yes & $\mathbf{<0.001}$ & Yes \\
XRay Vision UNet & $\mathbf{<0.001}$ & Yes & $\mathbf{<0.001}$ & Yes \\
UPerNet & $\mathbf{<0.001}$ & Yes & $\mathbf{<0.001}$ & Yes \\
UNet & $\mathbf{<0.001}$ & Yes & $\mathbf{<0.001}$ & Yes \\
ConvNeXt UPerNet & $\mathbf{<0.001}$ & Yes & $\mathbf{<0.001}$ & Yes \\
RepLKNet DeepLab & $\mathbf{<0.001}$ & Yes & $\mathbf{<0.001}$ & Yes \\
\hline
\end{tabular*}
\end{table*}

\section{Ethics statement}

The dataset presented in this work comprises exclusively retrospective, completely anonymized X-ray angiographic images obtained from routine clinical examinations. Prior to data access, all identifiable personal information was permanently stripped in accordance with institutional data protection policies. At the dataset release all additional demographic metadata was removed to prevent patient re-identification. Written informed consent was obtained from all subjects involved in the study.



\end{document}